\PassOptionsToPackage{table}{xcolor}
\documentclass{article}

\PassOptionsToPackage{numbers, compress}{natbib}

\usepackage[preprint]{neurips_2025}

\usepackage[utf8]{inputenc} 
\usepackage[T1]{fontenc}    
\usepackage{hyperref}       
\usepackage{url}            
\usepackage{booktabs}       
\usepackage{amsfonts}       
\usepackage{nicefrac}       
\usepackage{microtype}      

\usepackage{amsmath}
\usepackage{mathrsfs}
\usepackage{graphicx}
\usepackage{verbatim}
\usepackage{subfigure}
\usepackage{makecell,multirow}

\usepackage{float}
\usepackage{color}
\usepackage{bbm}
\usepackage[linesnumbered,ruled]{algorithm2e}
\usepackage{algorithmic}
\usepackage{multicol}
\usepackage{wrapfig}
\usepackage{bm}
\usepackage{array}
\usepackage{setspace}
\usepackage{enumitem}
\usepackage{soul}
\usepackage{minitoc}
\usepackage{caption}
\usepackage{amsthm}
\newtheorem{lemma}{Lemma}[section]
\newtheorem{theorem}{Theorem}[section]
\newtheorem{definition}{Definition}[section]

\definecolor{macaron1}{RGB}{221,241,214}
\definecolor{macaron2}{RGB}{244,220,224}
\definecolor{macaron3}{RGB}{210,233,241}
\definecolor{macaron4}{RGB}{231,214,226}

\definecolor{myblue}{RGB}{194, 214, 232}
\definecolor{myred}{RGB}{234, 185, 181}
\definecolor{mycolor}{RGB}{245, 223, 187}
\usepackage[table]{xcolor}         

\newboolean{useBlue}
\setboolean{useBlue}{False} 
\newcommand{\revise}[1]{%
  \ifthenelse{\boolean{useBlue}}%
    {\textcolor{blue}{#1}}%
    {\textcolor{black}{#1}}%
}

\newcommand{\figureresolution}{high resolution}

\newcommand{\ourmodel}{DF}

\title{Decision Flow Policy Optimization}

\author{%
  Jifeng Hu \\
  \And
  Sili Huang \\
  \And
  Siyuan Guo \\
  \And
  Zhaogeng Liu \\
  \AND
  Li Shen \\
  \And
  Lichao Sun \\
  \And
  Hechang Chen\thanks{Corresponding Author. Email address: chenhc@jlu.edu.cn (H. Chen) and yichang@jlu.edu.cn (Y. Chang). } \\
  \And
  $\text{Yi Chang}^{*}$  \\
  \And
  Dacheng Tao \\
}

\begin{document}

\maketitle

\begin{abstract}
    In recent years, generative models have shown remarkable capabilities across diverse fields, including images, videos, language, and decision-making. 
    By applying powerful generative models such as flow-based models to reinforcement learning, we can effectively model complex multi-modal action distributions and achieve superior robotic control in continuous action spaces, surpassing the limitations of single-modal action distributions with traditional Gaussian-based policies.
    Previous methods usually adopt the generative models as behavior models to fit state-conditioned action distributions from datasets, with policy optimization conducted separately through additional policies using value-based sample weighting or gradient-based updates. 
    However, this separation prevents the simultaneous optimization of multi-modal distribution fitting and policy improvement, ultimately hindering the training of models and degrading the performance. 
    To address this issue, we propose Decision Flow, a unified framework that integrates multi-modal action distribution modeling and policy optimization.
    Specifically, our method formulates the action generation procedure of flow-based models as a flow decision-making process, where each action generation step corresponds to one flow decision. 
    Consequently, our method seamlessly optimizes the flow policy while capturing multi-modal action distributions.
    We provide rigorous proofs of Decision Flow and validate the effectiveness through extensive experiments across dozens of offline RL environments.
    Compared with established offline RL baselines, the results demonstrate that our method achieves or matches the SOTA performance.
\end{abstract}

\section{Introduction}

Benefiting from the remarkable capability of generative models in modeling complex data distributions, flow-based models have achieved substantial progress in diverse fields, including image generation~\cite{esser2024scaling, dao2023flow, lipman2022flow, ren2024flowar}, video synthesis~\cite{jin2024pyramidal, zheng2024open, shrivastava2024video, liang2024flowvid}, language modeling~\cite{gat2024discrete, liu2023generative, davis2024fisher, sriram2024flowllm}, and robotic control~\cite{li2025generative, black2024pi_0, hu2024adaflow, yang2025policy}. 
In reinforcement learning (RL), many studies increasingly investigate leveraging the powerful expressivity of generative models, particularly flow-based models, to effectively master the capability of modeling multi-modal action distributions~\cite{zhang2025energy, park2025flow, zhang2024affordance}. 
This capability significantly overcomes the inherent limitations of traditional Gaussian-based policies, which can only represent unimodal action distributions~\cite{he2023diffusion, li2024learning}.

A naive approach to applying flow-based models in RL is employing the flow matching loss (refer to Section~\ref{Preliminary} for details) to imitate the state-conditional action distribution~\cite{rouxel2024flow, zheng2023guided}; we refer to policies trained in this manner as behavior flow policies. 
Compared to Gaussian-based policies, behavior flow policies can capture multi-modal action distributions, thereby enhancing policy expressivity~\cite{ada2024diffusion, wang2022diffusion}. 
However, this learning strategy is limited to merely mimicking the actions provided by the dataset, making it challenging to perform effective policy optimization, especially when learning from sub-optimal datasets~\cite{zhang2024preferred, ding2024diffusion, he2023diffusion}.

To optimize the actions generated by flow-based models, value functions are typically introduced to guide the action generation process, as demonstrated in previous studies~\cite{zhang2025energy, wang2022diffusion, mao2024diffusion}. 
However, the actions generated by the flow-based model involve multiple rollouts of ODE solvers, resulting in practical instability~\cite{park2025flow, black2023training}.
This is primarily because maximizing the Q-value-based objectives requires performing backpropagation along the multiple flow generation steps~\cite{gao2025behavior, hansen2023idql}. 
Consequently, existing studies commonly separate multi-modal action distribution modeling from policy optimization, thereby avoiding direct optimization of the flow policy~\cite{park2025flow, lu2023contrastive, kang2023efficient}.
The straightforward and effective approach is to treat the behavior flow policy as a reference policy and introduce an additional deterministic policy that generates actions in a single step to perform behavior flow constrained policy optimization~\cite{park2025flow, gao2025behavior}. 
Another representative research line employs value-weighted flow matching, which encourages the model to mimic action distributions with higher cumulative discounted returns by weighting the flow matching loss of different samples~\cite{ding2024diffusion, akimov2022let}. 
This method effectively facilitates learning optimal policies from sub-optimal datasets.
However, both approaches only optimize the final actions produced by the flow policy, making it challenging for action optimization signals to effectively propagate to intermediate steps in the action generation process, ultimately hindering the overall performance of the flow policy~\cite{gao2025behavior, lu2023contrastive}.

In this view, we propose \textbf{Decision Flow} (\ourmodel{}), a unified framework that integrates multi-modal action distribution modeling and policy optimization. In this paper, we formulate the generation process of flow models as a flow MDP, which is similar to the typical RL MDP, but with the difference of flow state, flow action, and flow reward shown in Definition~\ref{Flow MDP in mainbody}.
Through the flow reward definition, we can easily construct the optimization objectives with flow value functions for intermediate actions at any flow time step, where intermediate actions are the output vectors during the generation process of the flow policy.
Thus, the flow policy optimization is equal to maximizing the flow value of intermediate actions and minimizing the flow matching loss, which enables multi-modal action distributions modeling and flow policy optimization simultaneously.
We introduce two practical implementations of Decision Flow based on the specific assignment of flow MDP elements. 
Additionally, we provide rigorous proofs of the convergence of flow policy evaluation and improvement, where the theoretical results show that the flow value functions can instruct the generation to higher-return action regions.

To verify the effectiveness of our method, we select offline RL benchmarks D4RL with different difficulties and compare our method with dozens of established offline RL baselines, such as traditional offline RL methods, diffusion-based methods, transformer-based methods, and flow-based methods.
Totally, we compare with 30+ baselines on 20+ tasks.
Through the extensive experiments, we demonstrate that our method achieves and matches the SOTA performance in most environments.

\section{Related Work}

\noindent\textbf{Offline RL.}~~~Offline Reinforcement Learning (RL)~\cite{levine2020offline, kumar2020conservative, kostrikov2021offline, liu2024selfbc, light2024dataset, fujimoto2021minimalist, yu2021combo, ghosh2022offline, rezaeifar2022offline} has advanced through numerous algorithmic improvements that address distribution shift using conservative policies and constraints.
For example, Conservative Q-Learning (CQL)~\cite{kumar2020conservative} introduces lower-bound constraints on value functions during value iteration, reducing overestimation of unseen actions and achieving better performance in offline control tasks like robotics manipulation and Atari games. 
TD3+BC~\cite{fujimoto2021minimalist} proposes to enhance the performance of TD3 with behavioral cloning loss, effectively balancing value iteration and behavior policy constraints for robust offline RL performance.
Recently, policy improvement through implicit constraints emerged as another effective approach, for example, Implicit Q-Learning (IQL)~\cite{kostrikov2021offline} uses expectile regression to approximate advantage functions and derives policies through weighted behavioral cloning, achieving state-of-the-art results on D4RL benchmarks without evaluating out-of-dataset actions.
Besides, Model-based methods propose using the fictitious data generated from the task's world model to reduce distribution shift.
For example, Conservative Model-based Policy Optimization (COMBO)~\cite{yu2021combo}, as one representative model-based method, incorporates environment models into offline training with value penalties on generated data, achieving robust performance without explicit uncertainty estimation. 
Ensemble-based approaches~\cite{an2021uncertainty, song2023ensemble, lou2024uncertainty} have also proven effective at measuring uncertainty. 
As an example, the Uncertainty-Driven Offline RL algorithm (EDAC)~\cite{an2021uncertainty} employs truncated averaging across multiple Q functions to penalize high-variance estimates. 
By simply increasing the number of Q-networks, EDAC achieves better results across most offline benchmark tasks through its diverse Q-value ensembles.
Besides, generative models~\cite{franceschelli2024reinforcement, cao2023reinforcement}, such as transformer-based, diffusion-based, and flow-based models, are increasingly becoming a powerful method to realize elaborate control in RL.

\noindent\textbf{Transformer RL.}~~~
Recent research has explored incorporating transformer architectures from sequence modeling into offline RL to leverage long-term dependencies in offline data~\cite{chen2021decision, janner2021offline, zhang2023saformer, hu2023instructed, furuta2021generalized, wang2022bootstrapped, hu2022transforming}. 
A notable example is the Decision Transformer~\cite{chen2021decision}, which frames RL as conditional sequence modeling with transformers. 
Without explicit value functions, it achieves competitive performance on various offline benchmarks. 
Similarly, Trajectory Transformer (TT)~\cite{janner2021offline} treats complete trajectories as sequences and uses beam search for action planning.
Furthermore, TT is extended to multi-task scenarios with Multi-game Decision Transformer, which demonstrated that a single shared-parameter model could master 46 Atari games at near-human levels through offline training. 
Transformer models have proven effective in offline RL by leveraging sequential dependencies and rich trajectory data for decision-making~\cite{hu2022transforming, hu2023instructed}. 
However, studies indicate that return-conditioned transformer policies may be unstable in highly stochastic environments, suggesting the need for value-guided mechanisms to enhance robustness~\cite{chen2024deep}.

\noindent\textbf{Diffusion RL.}~~~
Diffusion Models~\cite{janner2022planning, fontanesi2019reinforcement, ajay2022conditional, gao2025behavior, chi2023diffusion, li2023efficient, ding2024diffusion} have been introduced into offline RL to handle multi-modal policies and long-horizon planning by leveraging their capability in modeling complex distributions. 
The Diffuser~\cite{janner2022planning} framework formulates decision-making as trajectory denoising: it first trains a diffusion model to generate complete trajectories, then uses classifier guidance to steer sampling toward high-return trajectories. 
In robotics control tasks~\cite{chi2023diffusion}, diffusion-based planning methods like Diffuser achieve comparable performance to traditional approaches while offering enhanced multi-modal behavior expression. 
The recent proposed diffusion policy~\cite{wang2022diffusion, gao2025behavior, he2023diffusion} treats robot policies as conditional diffusion processes, iteratively refining actions through denoising, where the key advantages of diffusion policies lie in their natural representation of multi-modal action distributions and stable training in high-dimensional action spaces. 
Furthermore, diffusion models have been utilized for offline data augmentation and trajectory stitching~\cite{li2024diffstitch}, enhancing learning capabilities in sparse-reward tasks~\cite{kang2023efficient}. 
While diffusion-based methods show promise in handling complex trajectory distributions and multi-modal policies in robotics and game AI, their inference speed remains a challenge due to iterative sampling, prompting ongoing research to optimize computational efficiency for real-time decision-making~\cite{huang2024diffusion}.

\noindent\textbf{Flow Matching RL.}~~~
Flow-based methods~\cite{park2025flow, akimov2022let, yang2023flow, urain2020imitationflow, papamakarios2021normalizing, zhang2025energy, zheng2023guided} have demonstrated superior performance in robot navigation tasks like locomotion compared to direct policy learning in original action spaces. 
Normalizing Flow models, which enable invertible mappings between data distributions and latent spaces, have been applied to enhance offline RL policy learning in two main aspects: action space modeling and hierarchical policy abstraction~\cite{papamakarios2021normalizing}. 
For addressing distribution mismatch between behavior and learning policies in offline RL, \citet{akimov2022let} first learn a latent action space encoder using normalizing flows on offline data, mapping complex action distributions to learnable low-dimensional representations, followed by policy network training in the latent domain. 
This approach inherently ensures conservatism without additional value penalties by constraining policies to generate only latent actions within the data support. 
Recent studies~\cite{park2025flow, urain2020imitationflow} investigate optimizing the flow policy by performing value-based gradient guided on an additional one-step flow model or by weighting the flow matching loss with corresponding energies. 
Additionally, flow models can also enable skill extraction and decomposition in offline RL~\cite{schroecker2020universal}.
Overall, flow-based methods leverage invertible distribution modeling capabilities to enhance the representation and utilization of complex continuous action distributions in offline RL, particularly benefiting robotics control by avoiding out-of-distribution actions and effectively utilizing demonstration data~\cite{zheng2023guided}.

\section{Preliminary}\label{Preliminary}
\noindent\textbf{Classical RL.}~~~Usually, the reinforcement learning (RL) problem is formulated as a Markov decision process (MDP), that is defined by the tuple $\mathcal{M}=\langle\mathcal{S},\mathcal{A},\mathcal{P},r,\gamma\rangle$, where $\mathcal{S}$ denotes the state space, $\mathcal{A}$ is the action space, $\mathcal{P}(s^\prime |s,a)$ is the state transition probability function, $r(s,a)$ represents the reward function, and $\gamma$ is the discount factor.
Typically, finding the optimal policy $\pi^*$ that maximizes the expected cumulative discounted reward can be formulated as the following objective
\begin{equation*}
    \pi^*=\arg \max_{\pi} \mathbb{E}_\pi[\sum_{n=0}^{\infty}\gamma^nr(s^n,a^n)],
\end{equation*}
where we use superscript $n$ to represent the RL time step.

\noindent\textbf{Flow RL.}~~~Traditional RL methods require only one time forward calculation to output actions. 
In contrast, flow models require multiple forward passes in one generation process to produce action vectors. 
Flow models try to map state-dependent action distributions to simple distributions (e.g., Gaussian distributions) through invertible probability paths, enabling gradual sampling from multi-modal action distributions.
Suppose that there exists a probability path that supports the transformation from simple distribution $p_0(x_0)$ to complex data distribution $p_1(x_1)$, where we usually define the simple distribution as Gaussian distribution, i.e., $x_0~\sim\mathcal{N}(0,\bm{I})$.
The intermediate distribution $p_t(x_t)$ on the probability path can be calculated through a function $\phi_t(x)$.
Generation process corresponds to learning the velocity field of $\phi_t(x)$ w.r.t. flow time step $t$:
\begin{equation}\label{velocity field}
    \frac{dx_t}{dt}=\frac{d\phi_t(x)}{dt}=u(\phi_t(x))=u(x_t,t),
\end{equation}
where we use $u(\phi_t(x))$ or $u(x_t,t)$ to denote the velocity field at certain intermediate point $x_t$ and flow time step t.
Generation from the flow model $u_\theta$ can be expressed as the following general formula:
\begin{equation}\label{ODE STEP transition}
    x_{t+\Delta t}=ODEStep(u_\theta(x_t,t), x_t),
\end{equation}
where $\theta$ is the parameter of flow model $u_{\theta}$, $x_{t+\Delta t}$ denotes the rollout results with $u_{\theta}$, and usually we select Euler method as $ODEStep(.)$~\cite{zheng2023guided}, i.e., $x_{t+\Delta t}=x_t+\Delta t*u_\theta(x_t,t)$.
In practice, we need to define the formula of $\phi_t(x)$, such as the most widely-used definition: $\phi_t(x|x_1)=x_t=(1-t)x_0+tx_1$, where we specify linear transformation from data $x_1$ to simple distribution $x_0$.

Thus, the flow matching loss is given by 
\begin{equation}
    \mathcal{L}_{fm}=\mathbb{E}_{x_t\sim p_t(x_t),t\sim U(0,1)}[||u_{\theta}(x_t,t)-u(x_t,t)||^2_2],
\end{equation}
where $t$ is sampled from the uniform distribution.
However, the above objective is intractable to optimize because 1) $u(x_t,t)$ is a complicated object governing the joint transformation between two high-dimensional distributions; 2) we can not obtain the data that comes from $p_t(x)$.
Fortunately, the objective can be simplified drastically when conditioning on a single target example $x_1$, which leads to the tractable conditional flow matching objective
\begin{equation}\label{conditional flow matching loss}
    \mathcal{L}_{cfm}=\mathbb{E}_{x_1\sim p_1(x_1),x_t\sim p_t(x_t|x_1),t\sim U(0,1)}\left[||u_\theta(x_t,t)-u(x_t|x_1,t))||_2^2\right],
\end{equation}
where we can leverage $\phi_t(x|x_1)$ to simultaneously sample $x_1$ and $x_t$, which leads to the conditional velocity field $u(x_t|x_1,t)$, that can be used to train the above loss.
In Appendix~\ref{Simplification of Flow Matching}, we can prove that $\mathcal{L}_{fm}$ and $\mathcal{L}_{cfm}$ are equal and provide the same gradient of $u_\theta$~\cite{lipman2024flow}.
In order to adapt the flow model to RL, we just need to replace $x_1$, $x_0$, and $x_t$ with $a_1|s$, $a_0|s$, and $a_t|s$, respectively.
Then we can generate state-conditional actions from the simple Gaussian distribution.

\section{Flow as RL}\label{Flow as RL}

\subsection{Flow Reinforcement Learning}

\begin{definition}\label{Flow MDP in mainbody}
(Flow MDP)
We regard the generation process induced by the velocity field as Flow-MDP $\langle\mathcal{S}^f,\mathcal{A}^f,\mathcal{P}^f,r^f,\gamma^f\rangle$, where the flow state at the flow time step t is denoted as $s^f_t\in\mathcal{S}^f$, the flow action is defined as $a^f_t\in\mathcal{A}^f$, and $\mathcal{P}^f$ is deterministic (induced by Equation~\eqref{ODE STEP transition}) for flow MDP.
$t\in[0,1]$ is continuous for the flow model, while in practice, for generation simplification, we discretize the range $[0, 1]$ into $T$ discrete flow time steps for each action generation.
Thus, we introduce discrete flow time step $\tau\in\{0,..., T\}$, and we have the transformation between $t$ and $\tau$: $\Delta t=\frac{1}{T}$, $t=\tau*\Delta t$.
$a_t$ and $a_\tau$ are just two different expressions with continuous $t$ and discrete $\tau$. 
The reward is defined as
\begin{equation}\label{flow mdp reward definition in mainbody}
    r_\tau^{flow}=
        \begin{cases}
            r^f_\tau, &0\leq \tau < T,\\
            Q(s^n,a_1^n), &\tau=T,
    \end{cases}
\end{equation}
where $Q$ is the traditional RL state-action value function, $n$ denotes RL time step, $a_1$ is the final output of the flow policy, and we slightly abuse $a_1$ and $a$ because they all represent the actions that can be used to interact with the environment.
However, we should note that the intermediate actions $a_t$ and $a_\tau$ are different from $a_1$ and $a$.
We use $\gamma^f$ to represent the discounted factor for flow MDP.
Besides, we use $\mathcal{T}^f$ to denote the Bellman transition. 
\end{definition}

By formulating the generation process of flow as a decision process, we introduce the Flow MDP (Definition~\ref{Flow MDP in mainbody}) that is similar to the traditional RL MDP.
Based on this definition, we introduce the Decision Flow method below, where we specify the quantity of the Flow MDP tuple and introduce two practical implementations.

\subsection{Direction-Oriented Decision Flow}\label{Direction-Oriented Decision Flow}

In this section, we introduce Direction-Oriented Decision Flow (\ourmodel{}-dir), which directly aligns the expected return of intermediate actions with the final action values $Q$. 
Unless otherwise specified, when we refer to `action,' we mean the RL action.
The `flow action' and `intermediate action' refer to the flow RL expression.
Below, we specify the elements of flow RL as follows:
\begin{align}
    &s^f=[s,a_t]; a^f=\frac{da_t}{dt}=u_\theta(s, a_t, t),\\
    &r_\tau^f=0; \gamma_\tau^f=1, \forall \tau\in[0,1,...,T],\\
    &Q^f_{\Psi}(s^f, a^f)=Q^f(s,a_t,u_\theta(s, a_t, t)),\\
    &V^f_{\Omega}(s^f)=V^f(s,a_t,\hat{u}_\theta(s, a_t, t)),
\end{align}
where $s^f=[s,a_t]$ and $a^f=u_\theta(s, a_t, t)$ represent the specific flow state and flow action, $u_\theta(s, a_t, t)$ represents the velocity field with magnitude and direction, $\hat{u}_\theta(s, a_t, t)=\frac{u_\theta(s, a_t, t)}{||u_\theta(s, a_t, t)||}$ denotes the normalized velocity field that only has direction. 
We use the flow action with direction and magnitude to define the flow action value function $Q^f$, and the normalized flow action just with direction to define the flow direction value function $V^f$.
$Q^f_{\Psi}(s^f,a^f)$ and $V^f_{\Omega}(s^f)$ are the parameterized flow action value function and flow direction value function, $\Psi$ and $\Omega$ are the corresponding parameters.
Obviously, by formulating the flow generation process as a reinforcement learning problem, we can use the flow value functions to instruct the flow policy to generate a better flow decision, i.e., the intermediate velocity that points to a higher return action region.
Apart from the above specification, we also need to introduce the traditional $Q$ function that implies Bellman error or other methods to learn the expected return prediction of final output actions from flow policy~\cite{van2016deep, kostrikov2021offline}.
Through the $Q$ function, we introduce the training objectives of flow value functions in Lemma~\ref{Critic Consistency in mainbody} and show that $Q^f_{\Psi}$ and $V_{\Omega}^f$ will converge as $Q\rightarrow Q^*$, and we postpone the detailed proofs in Appendix~\ref{Proof of Direction-Oriented Decision Flow}.

\begin{lemma}\label{Critic Consistency in mainbody}
(Critic Consistency)
If $Q\rightarrow Q^*$, where $Q^*$ is the optimal conventional critic and $Q^f$ and $V^f$ with sufficient model capacity, and the objectives $\mathcal{L}_{Q^f}$ and $\mathcal{L}_{V^f}$ is defined as
\begin{align}\label{DF-dir flow Q loss}
    \mathcal{L}_{Q^f}&=\mathbb{E}\left[||Q^f_{\Psi}-Q||_2^2\right],\\ \label{DF-dir flow V loss}
    \mathcal{L}_{V^f}&=\mathbb{E}\left[||V^f_{\Omega}-Q||_2^2\right].
\end{align}
Then, we will conclude that
\begin{equation}
    Q^{f*}(s,a_t,u_t)=V^{f*}(s,a_t,\hat{u}_t)=Q^*(s,a_1),\forall t,a_t.
\end{equation}
\end{lemma}

As the convergence of flow value functions, we show that the updating signals from $Q^f$ and $V^f$ are aligned with the traditional $Q$ function, and the intermediate flow actions (velocity field) will lead to the high-return region.
Mathematically, the conclusions are shown in Lemma~\ref{Direction Optimality in mainbody} and Lemma~\ref{Monotone Critic Improvement in mainbody}.
Note that the proof of Lemma~\ref{Direction Optimality in mainbody} is also suitable for $Q^{f*}$, because $Q^{f*}$ and $V^{f*}$ only have a difference in the velocity magnitude, which does not change the results shown in Appendix~\ref{Proof of Direction-Oriented Decision Flow}.  

\begin{lemma}\label{Direction Optimality in mainbody}
(Direction Optimality)
Under the results of Lemma~\ref{Critic Consistency in mainbody}, the gradient of $V^{f*}$ with respect to $\hat{u}_t$ vanishes if and only if $\hat{u}_t$ is aligned with the gradient of action $\nabla_{a_t}Q^*(s, a_t)$.
\end{lemma}

\begin{lemma}\label{Monotone Critic Improvement in mainbody}
(Monotone Critic Improvement)
Let $a_{t+\Delta t}=a_t+h\lambda\hat{u}_t$, where $\hat{u}$ satisfies the Lemma~\ref{Direction Optimality in mainbody} for small $h\lambda<\epsilon$. 
Then
\begin{equation}
    Q^*(s,a_{t+\Delta t}) \geq Q^*(s,a_t),
\end{equation}
where the equality holds only when $\nabla_{a_1} Q^*(s,a_1)=0$.
\end{lemma}

\begin{theorem}\label{flow policy convergence in mainbody}
Under the assumptions that 1) the traditional $Q$ function converges to the optimal value, 2) the flow value functions $Q^f$ and $V^f$ have sufficient capacity, and the training objectives are defined in Lemma~\ref{Critic Consistency in mainbody}, 3) the generation step size $h\lambda$ is small, 4) $Q$ is continuously differentiable over action space and attains a unique maximum $a^*_1=\arg\max_a Q(s,a)$ and Lemma~\ref{Critic Consistency in mainbody},~\ref{Direction Optimality in mainbody},and~\ref{Monotone Critic Improvement in mainbody}, we conclude that the generated actions by flow policy converge to the optimal actions $a^*_1$ and the flow policy $u_\theta$ is optimal.
\end{theorem}

In Theorem~\ref{flow policy convergence in mainbody}, we show that the trained flow policy $u_\theta$ will be improved as the convergence of flow $Q$ and $V$ functions.
Considering that the key function of the velocity field's magnitude and direction in flow policy evaluation, we named this method Direction-Oriented Decision Flow.
In practice, we select offline RL tasks to verify the effectiveness of this method.
We also introduce an additional policy constraint term in the training objective, as shown in the following problem
\begin{equation}\label{DF-dir policy loss}
    \begin{aligned}
        \min&~~\mathcal{L}_{u_\theta}^{\ourmodel{}-dir}=-\mathbb{E}_{s, a, t\sim U(0,1),a_t=\phi_t(a_t|a)}[Q^f_{\Psi}(s,a_t,u_\theta(s,a_t,t))-V^f_{\Omega}(s,a_t,\hat{u}_\theta(s,a_t,t))]\\
        &~~~~~~~~~~~~~~~~~~~+\rho*\mathcal{D}(u_\theta(s,a_t,t)||u_v(s,a_t,t)),
    \end{aligned}
\end{equation}
where $t$ is sampled from uniform distribution $U(0,1)$, $a_t$ can be obtained with $\phi_t(\cdot)$, $u_v$ is the behavior flow policy, $\mathcal{D}(u_{\theta}||u_v)=||u_\theta - u_\upsilon||_2$ is the divergence between $u_\theta$ and $u_v$, and $\rho$ is used as tradeoff between these two terms of the objective.

\subsection{Divergence-Oriented Decision Flow}\label{Divergence-Oriented Decision Flow}

Different from Direction-Oriented Decision Flow, where the flow value of every intermediate flow time step directly aligns with the final flow time step reward, i.e., $Q^f$ and $V^f$ align with $Q$, in this section, we introduce another method, Divergence-Oriented Multi-step Decision Flow (\ourmodel{}-div), that introduces intermediate flow step rewards defined with divergence.
Similarly, we first specify that
\begin{align}
    &s^f=[a^{n-1},s^n,a_t^n]; a^f=\frac{da_t}{dt}=u_\theta(s, a_t, t)\\
    &r_\tau^f=-\mathcal{D}(u_{\theta}(s,a_{\tau},\tau*\Delta t)||u_v(s,a_{\tau},\tau*\Delta t)), \gamma_\tau^f=1, \forall \tau\in[0,1,...,T],\\
    &V^f_{\chi}(s_f)=V^f(a^{n-1},s^n,a_t^n),
\end{align}
where $\mathcal{D}(u_\theta||u_v)=||u_\theta - u_\upsilon||_2$.
Note that we only introduce one flow value function $V^f_{\chi}(s_f)$ parameterized by $\chi$ to model the expected return of every flow time step.

This specification formulates the flow-based RL as a nested MDP where the generation process of the flow model is the inner MDP and the RL transition is the outer MDP, we define the final step reward of the flow MDP as $Q(s,a_1)$, and the intermediate reward is defined as negative divergence between the flow policy and the behavior flow policy. 
Mathematically, the multi-step flow decision policy optimization problem is defined as $\max \mathbb{E}\left[\sum_{\tau=0}^{T}(\gamma^f_\tau)^\tau r_\tau^{flow}\right]$.
The corresponding flow MDP transition is defined as
\begin{align}
    \mathcal{T}_V^f V^f_{\chi}(a^{n-1},s^n,a_1^{n})&=Q(s,a_1),\\
    \mathcal{T}_V^f V^f_{\chi}(a^{n-1},s^n,a_t^{n})&=-\mathcal{D}(u_\theta||u_\upsilon)+V^f_{\chi}(a^{n-1},s^n,a_{t+\Delta t}^{n}),\label{multi-step MDP process intermediate time step}
\end{align}
where $\mathcal{T}_V^f V_{\chi}(a^{n-1},s^n,a_t^{n})$ represents the Bellman operator on the intermediate flow steps, $V^f_{\chi}$ is the flow value function, $a^{n-1}$ is the actions at the last time step, which facilitates stable training of $V^f_{\chi}$ (Refer to Section~\ref{Ablation Study} for more details.).

The optimal flow value function of the flow MDP satisfies
\begin{equation}
    V^{f*}_{\chi}(a^{n-1},s^n,a_t^n)=-\mathcal{D}(u_{\theta}||u_v)+V^{f*}_{\chi}(a^{n-1},s^n,a_{t+\Delta t}^n), 
\end{equation}
where the terminal condition $V^{f*}_{\chi}(a^{n-1},s^n,a_1^n)=Q^*(s^n,a_1^n)$.
Mathematically, inspired by the definition of traditional Bellman error~\cite{sutton1998reinforcement}, training the flow value function $V_\chi^f$ is equal to minimizing  
\begin{equation}\label{mse loss of DF-div in mainbody}
    \mathcal{L}_{V_\chi}^{f}=\frac{1}{2}\mathbb{E}_{a^{n-1},s^n,a^{n}, t\sim U(0,1),a^{n}_t=\phi_t(a^{n}_t|a^{n})}\left[V_{\chi}^f(a^{n-1},s^n,a_t^n)-\hat{g}_{V^f}\right],
\end{equation}
where 
\begin{equation*}
    \hat{g}_{V^f}=-\sum_{\tau=t/\Delta t}^{T-1}\mathcal{D}(u_{\theta}(s,a_{\tau*\Delta t}, \tau*\Delta t)||u_v(s,a_{\tau*\Delta t}, \tau*\Delta t))+Q(s,a_1)
\end{equation*}
is the Monte Carlo estimation of $V^{f*}_{\chi}(a^{n-1},s^n,a_t^n)$.
In practice, we can use flow policy to generate the whole trajectories of flow MDP and train the flow value function.

In order to improve the flow policy, flow policy evaluation is needed to evaluate the expected value at any flow time step $t$.
Following Equation~\eqref{mse loss of DF-div in mainbody}, we can train the $\chi$-parameterized flow value function $V^f_\chi$ that can be used to optimize the flow decision, i.e., velocity field output.
We postpone the proof in Lemma~\ref{Flow Policy Evaluation}.
In Lemma~\ref{Flow policy Improvement}, we prove that the flow policy can be improved with the flow value function $V^f_\chi$, which, as shown in Theorem~\ref{Flow Policy Iteration in mainbody}, leads to the iterative optimization of the flow value function and flow policy.

\begin{theorem}\label{Flow Policy Iteration in mainbody}
(Flow Policy Iteration)
Repeated application of flow policy evaluation (Lemma~\ref{Flow Policy Evaluation}) and flow policy improvement (Lemma~\ref{Flow policy Improvement}) to any flow policy $u\in\mathcal{U}$ converges to a flow policy $u_\theta^*$ such that $\mathcal{J}_{u^*_\theta}(a_t^n)\geq\mathcal{J}_{u_\theta}(a_t^n)$ for all $u\in\mathcal{U}$ and $(a^{n-1},s^n,a^n_t)\in{\mathcal{A}\times\mathcal{S}\times\mathcal{A}}$.
\end{theorem}

There is still a problem about flow policy training in practice, because in Equation~\eqref{mse loss of DF-div in mainbody} we can see that $a_t^n$ in $V^f_\chi$ comes from sampling rather than flow policy, which will lead to no gradient backpropagation from flow value function to flow policy.
In order to train the flow policy $u_\theta$, we need to perform one-step ODE flow $a_{t+\Delta t}=ODEStep(u_\theta(s,a_t,t), a_t)$ that ensures the gradient from $V^{f}_{\chi}$ to $u_\theta$.
Then, updating the flow policy $u_{\theta}$ is equivalent to minimizing the negative flow value function under the flow time step $t+\Delta t$ with the divergence constraint
\begin{equation}\label{DF-div policy loss}
    \min~~~\mathcal{L}^{DF-div}_{u_\theta}= \mathbb{E}_{a^{n-1},s^n,a^{n}, t}\left[\mathcal{D}(u_\theta(s^n,a^n_t,t)||u_\upsilon(s^n,a^n_t,t))-V^{f}_{\chi}(a^{n-1},s^n,a^n_{t+\Delta t})\right].
\end{equation}

\subsection{Discussion of \ourmodel{}-dir and \ourmodel{}-div}

Comparing the two decision flow methods, both DF-dir and DF-div regard the flow model's intermediate velocity outputs as decisions and focus on modeling and optimizing flow value functions at intermediate steps. 
These value functions guide flow policy optimization during intermediate flow steps, which constitutes the key difference from previous methods. 
In addition, we provide rigorous proofs in Appendix~\ref{Proof of Direction-Oriented Decision Flow} and ~\ref{Proof of Divergence-Oriented Multi-step Decision Flow} to establish the theoretical foundations of our proposed methods.
Besides, these two methods both encourage the optimization of flow policy under every intermediate flow step constraint from the behavior flow policy.

The key distinction between DF-dir and DF-div lies in how their value function are aligned. 
DF-dir directly aligns the flow value functions with the Q values that are calculated with the final generated actions, meaning that the policy evaluation at every intermediate flow step explicitly considers the quality of the final output actions.
In contrast, DF-div formulates the flow-based RL as a nested MDP, where flow MDP acts as the inner MDP and RL MDP serves as the outer MDP.
In this setting, flow policy evaluation at the intermediate flow step is no longer directly aligned with the Q values on the final actions; instead, the flow values incorporate the policy divergence as the intermediate flow step reward, thus leading to divergence-oriented flow policy optimization (Equation~\ref{DF-div policy loss}).

\section{Experiment}\label{Experiment}

\subsection{Environment Settings}\label{Environment Settings}

As is well known, D4RL~\cite{fu2020d4rl} is a standard and widely used benchmark for comparing model performance~\cite{kumar2020conservative, fujimoto2021minimalist, yang2022behavior, he2023diffusion, wang2024prioritized, park2025flow}. 
This benchmark includes a variety of robotic control tasks, such as HalfCheetah, Walker2d, and Hopper, and for each task, datasets of different quality levels, such as expert and medium, are provided. 
In our evaluation, we select three major categories of environments from D4RL: Gym-MuJoCo, Pointmaze, and Adroit. 
Specifically, Gym-MuJoCo includes three tasks (HalfCheetah, Walker2d, and Hopper), each associated with three dataset quality levels—medium (m), medium-replay (mr), and medium-expert (me)—resulting in a total of 9 task settings. 
According to the size of the maze layouts, Pointmaze provides three difficulty levels—umaze (u), medium (m), and large (l)—and two reward settings, leading to 6 task settings~\cite{fujimoto2019off}. 
Adroit contains 12 high-dimensional robotic manipulation task settings, where the datasets are collected under three types of conditions: human (h), expert (e), and cloned (c)~\cite{rajeswaran2017learning}.

\begin{table*}[t!]
\centering
\vspace{-1em}
\small
\caption{Offline RL algorithms comparison on D4RL Gym-MuJoCo environments. The baselines are composed of \colorbox{macaron1}{traditional RL methods}, \colorbox{macaron2}{transformer-based methods}, \colorbox{macaron3}{diffusion-based methods}, and \colorbox{macaron4}{flow-based methods}, where we use different colors to show the method's category.}
\label{Offline RL algorithms comparison on D4RL Gym-MuJoCo}
\resizebox{\textwidth}{!}{
\begin{tabular}{l | r r r | r r r | r r r | r }
\toprule
\specialrule{0em}{1.5pt}{1.5pt}
\toprule
Dataset & \multicolumn{3}{c|}{Med-Expert} & \multicolumn{3}{c|}{Medium} & \multicolumn{3}{c|}{Med-Replay} & \multirow{2}{*}{\makecell[r]{mean\\score}}\\
\cline{1-10}
\rule{0pt}{2.5ex} Env & HalfCheetah & Hopper & Walker2d & HalfCheetah & Hopper & Walker2d & HalfCheetah & Hopper & Walker2d &  \\
\midrule[1pt]
\rowcolor{macaron1}
AWAC 
& 42.8 & 55.8 & 74.5 & 43.5 & 57.0 & 72.4 & 40.5 & 37.2 & 27.0 & 50.1\\
\rowcolor{macaron1}
BC
& 55.2 & 52.5 & 107.5 & 42.6 & 52.9 & 75.3 & 36.6 & 18.1 & 26.0 & 51.9\\
\rowcolor{macaron1}
MOPO
& 63.3
& 23.7
& 44.6
& 42.3
& 28.0
& 17.8
& 53.1
& 67.5
& 39.0
& 42.1\\
\rowcolor{macaron1}
MBOP
& 105.9 & 55.1 & 70.2 & 44.6 & 48.8 & 41.0 & 42.3 & 12.4 & 9.7 & 47.8 \\
\rowcolor{macaron1}
MOReL
& 53.3 & 108.7 & 95.6  & 42.1 & 95.4 & 77.8 & 40.2 & 93.6 & 49.8 & 72.9 \\
\rowcolor{macaron1}
TAP
& 91.8
& 105.5
& 107.4
& 45.0
& 63.4
& 64.9
& 40.8
& 87.3
& 66.8
& 74.8 \\
\rowcolor{macaron1}
BEAR
& 51.7 & 4.0 & 26.0 & 38.6 & 47.6 & 33.2 & 36.2 & 10.8 & 25.3 & 30.4 \\
\rowcolor{macaron1}
BCQ
& 64.7 & 100.9 & 57.5 & 40.7 & 54.5 & 53.1 & 38.2 & 33.1 & 15.0 & 50.9 \\
\rowcolor{macaron1}
CQL
& 62.4 & 98.7 & 111.0 & 44.4 & 58.0 & 79.2 & 46.2 & 48.6 & 26.7 & 63.9 \\
\rowcolor{macaron1}
TD3+BC
& 90.7 & 98.0 & 110.1 & 48.3 & 59.3 & 83.7 & 44.6 & 60.9 & 81.8 & 75.3 \\
\rowcolor{macaron1}
IQL
& 86.7 & 91.5 & 109.6 & 47.4 & 66.3 & 78.3 & 44.2 & 94.7 & 73.9 & 77.0 \\
\rowcolor{macaron1}
PBRL
& 92.3
& 110.8
& 110.1
& 57.9
& 75.3
& 89.6
& 45.1
& 100.6
& 77.7
& 84.4 \\
\midrule
\rowcolor{macaron2}
DT
& 90.7 & 98.0 & 110.1 & 42.6 & 67.6 & 74.0 & 36.6 & 82.7 & 66.6 & 74.3 \\
\rowcolor{macaron2}
TT
& 95.0 & 110.0 & 101.9 & 46.9 & 61.1 & 79.0 & 41.9 & 91.5 & 82.6 & 78.9 \\
\rowcolor{macaron2}
BooT
& 94.0 & 102.3 & 110.4 & 50.6 & 70.2 & 82.9 & 46.5 & 92.9 & 87.6 & 81.9 \\
\midrule
\rowcolor{macaron3}
SfBC
& 92.6
& 108.6
& 109.8
& 45.9
& 57.1
& 77.9
& 37.1
& 86.2
& 65.1
& 75.6 \\
\rowcolor{macaron3}
D-QL@1
& 94.8 & 100.6 & 108.9 & 47.8 & 64.1 & 82.0 & 44.0 & 63.1 & 75.4 & 75.6 \\
\rowcolor{macaron3}
Diffuser
& 88.9
& 103.3
& 106.9
& 42.8
& 74.3
& 79.6
& 37.7
& 93.6
& 70.6
& 77.5 \\
\rowcolor{macaron3}
DD
& 90.6
& 111.8
& 108.8
& 49.1
& 79.3
& 82.5
& 39.3
& 100.0
& 75.0
& 81.8 \\
\rowcolor{macaron3}
IDQL
& 95.9 & 108.6 & 112.7 & 51.0 & 65.4 & 82.5 & 45.9 & 92.1 & 85.1 & 82.1 \\
\rowcolor{macaron3}
HDMI
& 92.1
& 113.5
& 107.9
& 48.0
& 76.4
& 79.9
& 44.9
& 99.6
& 80.7
& 82.6 \\
\rowcolor{macaron3}
AdaptDiffuser
& 89.6
& 111.6
& 108.2
& 44.2
& 96.6
& 84.4
& 38.3
& 92.2
& 84.7
& 83.3 \\
\rowcolor{macaron3}
DiffuserLite
& 87.8
& 110.7
& 106.5
& 47.6
& 99.1
& 85.9
& 41.4
& 95.9
& 84.3
& 84.4 \\
\rowcolor{macaron3}
HD-DA
& 92.5
& 115.3
& 107.1
& 46.7
& 99.3
& 84.0
& 38.1
& 94.7
& 84.1
& 84.6 \\
\midrule
\rowcolor{macaron4}
FQL
& 86.1
& 21.1
& 13.3
& 60.0
& 25.1
& 9.3
& 53.1
& 28.4
& 6.9
& 33.7 \\
\rowcolor{macaron4}
Flow
& 97.0
& 105.0
& 94.0
& 49.0
& 84.0 
& 77.0
& 42.0 
& 89.0
& 78.0 
& 79.0  \\
\rowcolor{macaron4}
CNF
& 96.2
& 108.6
& 112.3
& 50.6
& 69.3
& 83.6
& 45.8
& 89.0
& 82.0
& 81.9  \\
\midrule
\rowcolor{macaron4}
\ourmodel{}-dir
& 87.6\tiny{$\pm$3.0}
& 112.6\tiny{$\pm$1.6}
& 111.3\tiny{$\pm$0.8}
& 49.3\tiny{$\pm$0.9}
& 97.1\tiny{$\pm$6.1} 
& 87.2\tiny{$\pm$1.7}
& 44.2\tiny{$\pm$1.6} 
& 102.3\tiny{$\pm$1.1}
& 86.6\tiny{$\pm$13.6} 
& 86.5 \\
\rowcolor{macaron4}
\ourmodel{}-div
& 87.7\tiny{$\pm$1.8}
& 112.9\tiny{$\pm$2.0}
& 107.8\tiny{$\pm$1.1}
& 43.5\tiny{$\pm$1.1}
& 99.5\tiny{$\pm$0.7} 
& 83.9\tiny{$\pm$4.5}
& 41.5\tiny{$\pm$1.1} 
& 98.6\tiny{$\pm$2.3}
& 84.2\tiny{$\pm$3.2}  
& 84.4
\\
\bottomrule
\specialrule{0em}{1.5pt}{1.5pt}
\bottomrule
\end{tabular}}
\vspace{-0.5cm}
\end{table*}

\subsection{Performance Measurements}\label{Performance Measurements}

Maintaining the comparison with previous methods~\cite{kumar2020conservative, janner2022planning, fujimoto2021minimalist, chen2021decision}, we adopt the normalized score value as the performance metric because the normalized score can avoid the interference of various reward structures across different environments.
The score values of each baseline are calculated with $\mathcal{R}_{norm}=\frac{\mathcal{R}-\mathcal{R}_{random}}{\mathcal{R}_{expert}-\mathcal{R}_{random}}*100$, where $\mathcal{R}_{norm}$ is the normalized score, $\mathcal{R}_{random}$ denotes the episode return of random policy, and $\mathcal{R}_{expert}$ is the episode return of expert policy.

\subsection{Baselines Selection}\label{Baselines Selection}

As shown below, the baselines are classified into several categories.
In summary, we compared more than 30 competitive methods across over 20+ offline RL tasks.
Traditional RL methods, which include actor-critic algorithms such as AWAC~\cite{nair2020awac}, SAC~\cite{haarnoja2018soft}, IQL~\cite{kostrikov2021offline}, and BC, model-based algorithms, such as TAP~\cite{jiang2022efficient}, MOReL~\cite{kidambi2020morel}, MOPO~\cite{yu2020mopo}, and MBOP~\cite{argenson2020model}, policy-constraint algorithms, such as CQL~\cite{kumar2020conservative}, TD3+BC~\cite{fujimoto2021minimalist}, BCQ~\cite{fujimoto2019off}, PBRL~\cite{bai2022pessimistic}, and BEAR~\cite{kumar2019stabilizing}.
Transformer-based RL methods, including BooT~\cite{wang2022bootstrapped}, TT~\cite{janner2021offline}, and DT~\cite{chen2021decision}, etc.
Diffusion-based RL methods are composed of DiffuserLite~\cite{dong2024diffuserlite}, HD-DA~\cite{chen2024simple}, IDQL~\cite{hansen2023idql}, AdaptDiffuser~\cite{liang2023adaptdiffuser}, HDMI~\cite{li2023hierarchical}, SfBC~\cite{chen2022offline}, DD~\cite{ajay2022conditional}, Diffuser~\cite{janner2022planning}, D-QL~\cite{wang2022diffusion}, etc.
Flow-based RL methods contain CNF~\cite{akimov2022let}, Flow~\cite{zheng2023guided}, FQL~\cite{park2025flow}, and the flow-based variants of AWAC~\cite{nair2020awac}, DQL~\cite{wang2022diffusion}, and IDQL~\cite{hansen2023idql}, i.e., FAWAC, FBRAC, and IFQL.

\subsection{Results Analysis}\label{Results Analysis}

To show the effectiveness of our method, we conduct abundant experiments on D4RL Gym-MuJoCo tasks, which contain 9 offline RL task settings.
The baselines include representative algorithms from 4 categories: traditional RL methods, transformer-based methods, diffusion-based methods, and flow-based methods.
We report the experimental results of D4RL Gym-MuJoCo in Table~\ref{Offline RL algorithms comparison on D4RL Gym-MuJoCo}.
From the results, we can see that our method achieves SOTA performance compared with the recent flow-based RL methods.
Additionally, compared with other baseline categories, our method also surpasses or matches the best performance.
To the best of our knowledge, we are the first to formulate the generation process of the flow model as a flow Markov decision process and introduce flow value functions to improve the flow policy.
The results show that through the formulation of flow MDP, our method will lead to better flow action, i.e., the velocity field that points to the higher return action region.
Considering that more experiments are better to further validate the superiority of our method, we conduct experiments on Adroit tasks, which is a hand manipulation benchmark and contains several sparse rewards and high-dimensional robotic manipulation tasks~\cite{rajeswaran2017learning}.
As shown in Table~\ref{Offline RL algorithms comparison on Adroit}, our method achieves an approximate 16\% performance gain compared with the best flow-based baselines.
Compared with baselines from other categories, our method still reaches the SOTA performance.
We postpone more experiments and discussion in Appendix~\ref{Additional Experiments}.

\subsection{Ablation Study}\label{Ablation Study}

\begin{figure}[h]
\vspace{-1em}
 \begin{center}
 \ifthenelse{\equal{\figureresolution}{low resolution}}
    {\includegraphics[angle=0,width=0.99\textwidth]{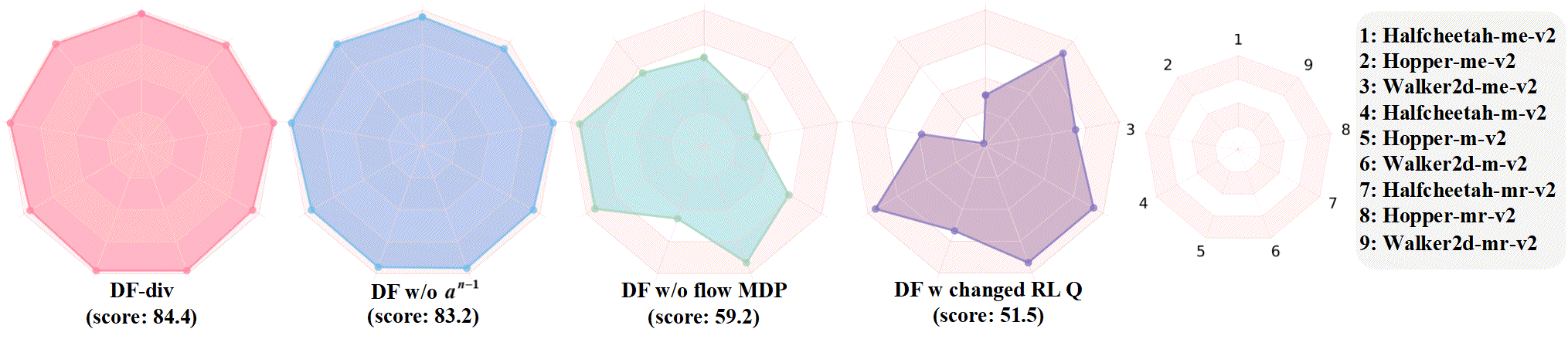}}
    {\includegraphics[angle=0,width=0.99\textwidth]{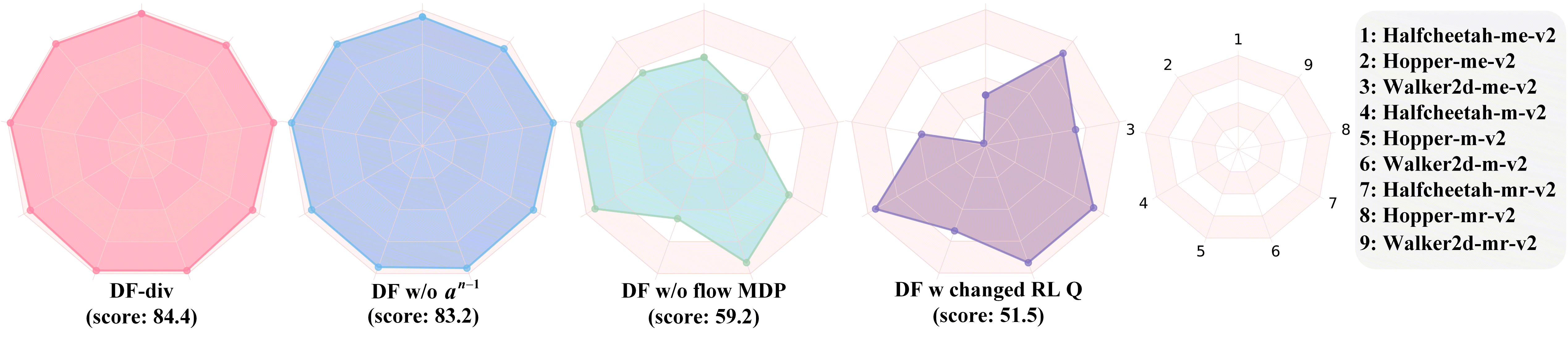}}
 \caption{Ablation study of Decision Flow. We investigate the importance of $a^{n-1}$, converting the flow as RL, flow value functions (i.e., flow critics) on the Gym-MuJoCo tasks. Each vertex represents one task, and the coordinates represent the evaluation results on the corresponding tasks. The average score across all tasks is shown under the name in each sub-figure.}
 \label{d4rl ablation}
 \end{center}
 \vspace{-0.3cm}
 \end{figure}

Firstly, we investigate the importance of $a^{n-1}$ used in \ourmodel{}-div, where the corresponding ablation method `\ourmodel{} w/o $a^n-1$' shows that adding more information $a^{n-1}$ into the flow value functions can further improve the performance.
When removing the flow MDP, i.e., `\ourmodel{} w/o flow MDP', then the flow policy can only mimic the action distribution from the dataset, which leads to inferior performance (59.2) compared with \ourmodel{}-div (84.4).
Additionally, we investigate other traditional Q estimation methods in `\ourmodel{} w changed RL Q' and report the overall performance in Figure~\ref{d4rl ablation}, where we use a variant of IQL~\cite{kostrikov2021offline} and CQL~\cite{kumar2020conservative} (Refer to Appendix~\ref{Training of Traditional Q Function} for more details). 
The results support the importance of the corresponding designs for our method.

\begin{table*}[t!]
\centering
\small
\caption{Offline RL algorithms comparison on Adroit, which contains 12 task settings for evaluation. The baselines are composed of \colorbox{macaron1}{traditional RL methods}, \colorbox{macaron2}{transformer-based methods}, \colorbox{macaron3}{diffusion-based methods}, and \colorbox{macaron4}{flow-based methods}, where we use different colors to show the method's category.}
\label{Offline RL algorithms comparison on Adroit}
\resizebox{\textwidth}{!}{
\begin{tabular}{l | r r r | r r r | r r r | r r r | r}
\toprule
\specialrule{0em}{1.5pt}{1.5pt}
\toprule
Task & \multicolumn{3}{c|}{pen} & \multicolumn{3}{c|}{hammer} & \multicolumn{3}{c|}{door} & \multicolumn{3}{c|}{relocate} & \multirow{2}{*}{\makecell[r]{mean\\score}} \\
\cline{1-13}
Dataset & human & expert & cloned & human & expert & cloned & human & expert & cloned & human & expert & cloned & \\
\midrule[1pt]
\rowcolor{macaron1}
BEAR 
& -1.0 & - & 26.5 & - & - & - & - & - & - & - & - & - & - \\
\rowcolor{macaron1}
BCQ 
& 68.9 & - & 44.0 & - & - & - & - & - & - & - & - & - & - \\
\rowcolor{macaron1}
CQL 
& 37.5 & 107.0 & 39.2 & 4.4 & 86.7 & 2.1 & 9.9 & 101.5 & 0.4 & 0.2 & 95.0 & -0.1 & 40.3 \\
\rowcolor{macaron1}
BC 
& 71.0 & 110.0 & 52.0 & 3.0 & 127.0 & 1.0 & 2.0 & 105.0 & 0.0 & 0.0 & 108.0 & 0.0 & 48.0 \\
\rowcolor{macaron1}
TAP 
& 76.5
& 127.4
& 57.4
& 1.4
& 127.6
& 1.2
& 8.8
& 104.8
& 11.7
& 0.2
& 105.8
& -0.2
& 51.9 \\
\rowcolor{macaron1}
IQL & 78.0 & 128.0 & 83.0 & 2.0 & 129.0 & 2.0 & 3.0 & 107.0 & 3.0 & 0.0 & 106.0 & 0.0 & 53.0 \\
\midrule
\rowcolor{macaron2}
DT 
& 79.5 & - & 75.8 & 3.7 & - & 3.0 & 14.8 & - & 16.3 & - & - & - & - \\
\rowcolor{macaron2}
GDT 
& 92.5 & - & 86.2 & 5.5 & - & 8.9 & 20.6 & - & 19.8 & - & - & - & - \\
\rowcolor{macaron2}
TT 
& 36.4 & 72.0 & 11.4 & 0.8 & 15.5 & 0.5 & 0.1 & 94.1 & -0.1 & 0.0 & 10.3 & -0.1 & 20.1 \\
\midrule
\rowcolor{macaron3}
TCD 
& 49.9 & 35.6 & 73.3 & - & - & - & - & - & - & 0.4 & 59.6 & 0.2 & - \\
\rowcolor{macaron3}
HD-DA & -2.6 & 107.9 & -2.7 & - & - & - & - & - & - & 0.0 & -0.1 & -0.2 & - \\
\rowcolor{macaron3}
DiffuserLite & 33.2 & 20.7 & 2.1 & - & - & - & - & - & - & 0.1 & 0.1 & -0.2 & - \\
\rowcolor{macaron3}
DD 
& 64.1
& 107.6
& 47.7
& 1.0
& 106.7
& 0.9
& 6.9
& 87.0
& 9.0
& 0.2
& 87.5
& -0.2
& 43.2 \\
\rowcolor{macaron3}
HDMI 
& 66.2
& 109.5
& 48.3
& 1.2
& 111.8
& 1.0
& 7.1
& 85.9
& 9.3
& 0.1
& 91.3
& -0.1
& 44.3 \\
\rowcolor{macaron3}
D-QL@1 
& 66.0
& 112.6
& 49.3
& 1.3
& 114.8
& 1.1
& 8.0
& 93.7
& 10.6
& 0.2
& 95.2
& -0.2
& 46.1 \\
\rowcolor{macaron3}
QGPO 
& 73.9
& 119.1
& 54.2
& 1.4
& 123.2
& 1.1
& 8.5
& 98.8
& 11.2
& 0.2
& 102.5
& -0.2
& 49.5 \\
\rowcolor{macaron3}
SRPO 
& 69.0
& 134.0
& 61.0
& 1.0
& 127.0
& 2.0
& 3.0
& 105.0
& 0.0
& 0.0
& 106.0
& 0.0
& 51.0 \\
\rowcolor{macaron3}
LD 
& 79.0
& 131.2
& 60.7
& 4.6
& 132.5
& 4.2
& 9.8
& 111.9
& 12.0
& 0.2
& 109.5
& -0.1
& 54.6 \\
\midrule
\rowcolor{macaron4}
FAWAC
& 67.0
& 118.0
& 62.0
& 2.0
& 118.0
& 1.0
& 2.0
& 103.0
& 0.0
& 0.0
& 105.0
& 0.0
& 48.0 \\
\rowcolor{macaron4}
FBRAC
& 77.0
& 119.0
& 67.0
& 2.0
& 119.0
& 2.0
& 4.0
& 104.0
& 0.0
& 0.0
& 105.0
& 1.0
& 50.0 \\
\rowcolor{macaron4}
IFQL
& 71.0
& 139.0
& 80.0
& 3.0
& 117.0
& 2.0
& 7.0
& 104.0
& 2.0
& 0.0
& 104.0
& 0.0
& 52.0 \\
\rowcolor{macaron4}
FQL
& 53.0
& 142.0
& 74.0
& 1.0
& 125.0
& 11.0
& 0.0
& 104.0
& 2.0
& 0.0
& 107.0
& 0.0
& 52.0 \\
\midrule
\rowcolor{macaron4}
\ourmodel{}-div
& 102.8
& 139.1
& 104.3
& 13.4
& 128.1
& 7.1
& 11.0
& 106.2
& 0.0
& 0.2
& 110.6
& 0.2
& 60.3 \\
\bottomrule
\specialrule{0em}{1.5pt}{1.5pt}
\bottomrule
\end{tabular}}
\vspace{-0.3cm}
\end{table*}

\section{Conclusion and Limitations}\label{Conclusion and Limitations}
In this paper, we propose Decision Flow, which formulates the flow process with the velocity field as a flow Markov decision process, thus integrating multi-modal action distribution modeling and policy optimization.
Additionally, we provide rigorous proofs of the improvements of flow value functions and the flow policy to prove the convergence of Decision Flow.
Finally, with systematic experiments, we demonstrate that our method achieves or matches the SOTA performance in dozens of tasks by comparing with 30+ baselines. 

For limitations, the rollout of multiple flow steps is one limitation for flow-based RL methods because one RL decision needs $T$ flow time steps in online tasks.
More research is needed to realize faster generation and better modeling with fewer flow-time steps.
Though the linear definition of $\phi_t(\cdot)$ (Equation~\eqref{velocity field}) is widely used in most flow-based methods and is good for action distribution modeling, we must emphasize that there are various probability paths worthy of study between $x_0$ and $x_1$.

\clearpage
\bibliography{reference}
\bibliographystyle{plainnat}

\clearpage
\appendix

\section{Pseudocode of Decision Flow}\label{Pseudocode}

\begin{algorithm}[h!]
\caption{Decision Flow (\ourmodel{}).}
\label{algorithm}
\begin{algorithmic}[1]
\STATE \textbf{Input:} Offline dataset $D$, training iteration $M$, environmental time limit $N$, generation steps $T$, discrete flow time step $\tau\in[0,T]$, continuous flow time step $t\in[0,1]$, $\Delta t=\frac{1}{T}$, flow policy $u_\theta$, pretrained behavior flow policy $u_v$, flow value functions $Q^f_{\Psi}$, $V^f_{\Omega}$, and $V^f_{\chi}$, traditional state-action value function $Q$
\STATE \textbf{Output:} Well-trained flow policy $u_\theta$
\STATE \textcolor{gray}{// Training}
\FOR{$i=1$ $\textbf{to}$ $M$}
    \STATE Train the traditional state-action value function $Q_\psi$ parameterized by $\psi$ with Equation~\eqref{IQL V loss}-~\eqref{IQL expectile loss}
    \IF{Flow Decision is \ourmodel{}-dir}
        \STATE \textcolor{gray}{// If we adopt the Direction-Oriented Decision Flow introduced in Section~\ref{Direction-Oriented Decision Flow}, the training process of the flow policy and flow value functions is as follows.}
        \STATE Train flow value functions $Q^f_{\Psi}$ and $V^f_{\Omega}$ with objectives $\mathcal{L}_{Q^f}$ and $\mathcal{L}_{V^f}$ (Equation~\eqref{DF-dir flow Q loss} and Equation~\eqref{DF-dir flow V loss})
        \STATE Improve flow policy $u_\theta$ by minimizing $\mathcal{L}_{u_\theta}^{\ourmodel{}-dir}$ (Equation~\eqref{DF-dir policy loss})
    \ELSIF{Flow Decision is \ourmodel{}-div}
        \STATE \textcolor{gray}{// If we adopt the Divergence-Oriented Decision Flow (Refer to Section~\ref{Divergence-Oriented Decision Flow} for introduction), the training process is as follows.}
        \STATE Train flow value function $V^f_{\Omega}$ with loss shown in Equation~\eqref{mse loss of DF-div in mainbody}
        \STATE Improve flow policy $u_\theta$ by minimizing $\mathcal{L}^{DF-div}_{u_\theta}$ (Equation~\eqref{DF-div policy loss})
    \ENDIF
\ENDFOR
\STATE // \textcolor{gray}{Evaluation}
\FOR{$n=1$ $\textbf{to}$ N}
    \STATE Receive state $s^n$ from the environment
    \STATE Sample $a^n_0$ from $\mathcal{N}(0,\bm{I})$
    \STATE Set $a^n_t=a^n_0$
    \FOR{$\tau=0$ $\textbf{to}$ $T-1$}
        \STATE Obtain the flow decision $u_\theta(s^n,a^n_{\tau*\Delta t}, \tau*\Delta t)$, i.e., the velocity field at $\tau*\Delta t$
        \STATE Perform one-step action generation with $u_\theta$ and $a^n_t$ (Equation~\eqref{ODE STEP transition}) and obtain $a^n_{t+\Delta t}$
        \STATE Set $a^n_t=a^n_{t+\Delta t}$
    \ENDFOR
    \STATE Interact with the environment with generated action $a^n_1$
\ENDFOR
\end{algorithmic}
\end{algorithm}

The training process and evaluation process of Decision Flow (\ourmodel{}) are shown in Algorithm~\ref{algorithm}.
In lines 3-15, we introduce the training of two variants of Decision Flow: Direction-Oriented Decision Flow and Divergence-Oriented Decision Flow.
During the evaluation (lines 16-27), we perform generation with Equation~\eqref{ODE STEP transition} for each state received from the environment, where the ODE step from $a_t$ to $a_{t+\Delta t}$ is the Euler method.
Noted that the flow time input of the flow policy is continuous values whose range is $[0,1]$, so we need to convert the discrete values $\tau$ to continuous values $\tau*\Delta t$.
After $T$ flow time steps, we obtain the generated action $a^n_1$ at RL time step $n$ to interact with the environment.

\section{Additional Experiments}\label{Additional Experiments}

\subsection{Experiments on Pointmaze}
Apart from the experiments introduced in Section~\ref{Results Analysis}, we also conduct the Pointmaze tasks with different reward settings.
Maze2D in Pointmaze is a 2D navigation task where the agent needs to match a specific location, and the goal is to learn the shortest path from the start point to the goal location with an offline dataset.
The trajectories are collected with waypoint sequences that are generated by a PD controller~\cite{fu2020d4rl}.
We compare our method with 19 baselines on the 6 task settings and report the results in Table~\ref{mazed2d and antmaze evaluation}.
The results show that our method reaches or matches the SOTA performance comparing with other baselines, which reveals the huge potential of flow-based RL methods.

\begin{table*}[t!]
\centering
\vspace{-1em}
\small
\caption{D4RL Pointmaze (maze2d) comparison. As introduced in Section~\ref{Environment Settings}, Pointmaze contains 6 offline RL settings according to the dataset qualities and reward settings. The baselines are composed of \colorbox{macaron1}{traditional RL methods}, \colorbox{macaron2}{transformer-based methods}, \colorbox{macaron3}{diffusion-based methods}, and \colorbox{macaron4}{flow-based methods}, where we use different colors to show the method's category.}
\label{mazed2d and antmaze evaluation}
\resizebox{0.9\textwidth}{!}{
\begin{tabular}{l | r r | r r | r r | r r r}
\toprule
\specialrule{0em}{1.5pt}{1.5pt}
\toprule
Environment & \multicolumn{2}{c|}{maze2d-umaze} & \multicolumn{2}{c|}{maze2d-medium} & \multicolumn{2}{c|}{maze2d-large} & \multirow{2}{*}{\makecell[c]{mean\\sparse\\score}} & \multirow{2}{*}{\makecell[c]{mean\\dense\\score}} & \multirow{2}{*}{\makecell[c]{mean\\score}}\\
\cline{1-7}
\specialrule{0em}{5.0pt}{1.5pt}
Environment type & sparse & dense & sparse & dense & sparse & dense &  & \\
\midrule
\specialrule{0em}{1.5pt}{1.5pt}
\rowcolor{macaron1}
 AWR
 & 1.0 & - & 7.6 & - & 23.7 & - & 10.8 & - & -\\
 \rowcolor{macaron1}
 BCQ     & 49.1 & - & 17.1 & - & 30.8 & - & 32.3 & - & -\\
 \rowcolor{macaron1}
 IQL     & 42.1 & - & 34.9 & - & 61.7 & - & 46.2 & - & -\\
 \rowcolor{macaron1}
 COMBO
 & 76.4 & - & 68.5 & - & 14.1 & - & 53.0 & - & -\\
 \rowcolor{macaron1}
 TD3+BC  & 14.8 & - & 62.1 & - & 88.6 & - & 55.2 & - & -\\
 \rowcolor{macaron1}
 BEAR    & 65.7 & - & 25.0 & - & 81.0 & - & 57.2 & - & -\\
  \rowcolor{macaron1}
 BC      & 88.9 & 14.6 & 38.3 & 16.3 & 1.5  & 17.1 & 42.9 & 16.0 & 29.5\\
 \rowcolor{macaron1}
 CQL     & 94.7 & 37.1 & 41.8 & 32.1 & 49.6 & 29.6 & 62.0 & 32.9 & 47.5\\
 \midrule
 \rowcolor{macaron2}
 DT      & 31.0  & - & 8.2 & - & 2.3 & - & 13.8 & - & -\\
 \rowcolor{macaron2}
 GDT
 & 50.4  & - & 7.8 & - & 0.7 & - & 19.6 & - & -\\
 \rowcolor{macaron2}
 QDT
 & 57.3  & - & 13.3  & - & 31.0  & - & 33.9 & - & -\\
 \rowcolor{macaron2}
 TT      & 68.7  & 46.6 & 34.9 & 52.7 & 27.6 & 56.6 & 43.7 & 52.0 & 47.9\\
 \midrule
 \rowcolor{macaron3}
 SfBC    
 & 73.9
 & - 
 & 73.8
 & - 
 & 74.4
 & -
 & 74.0 & - & - \\
 \rowcolor{macaron3}
 SynthER
 & 99.1
 & -
 & 66.4
 & -
 & 143.3
 & - 
 & 102.9 & - & -\\
 \rowcolor{macaron3}
 Diffuser& 113.9 & - & 121.5 & - & 123.0 & - & 119.5 & - & -\\
 \rowcolor{macaron3}
 HDMI
 & 120.1
 & - 
 & 121.8
 & -
 & 128.6
 & - 
 & 123.5 & - & -\\
 \rowcolor{macaron3}
 HD-DA   & 72.8  & 45.5 & 42.1 & 54.7 & 80.7 & 45.7 & 65.2 & 48.6 & 56.9\\
 \rowcolor{macaron3}
 TCD    & 128.1 & 29.8 & 132.9 & 41.4 & 146.4 & 75.5 & 135.8 & 48.9 & 92.4\\
 \rowcolor{macaron3}
 DD      & 116.2 & 83.2 & 122.3 & 78.2 & 125.9 & 23.0 & 121.5 & 61.5 & 91.5\\
 \midrule
 \rowcolor{macaron4}
 \ourmodel{}-div& 104.0 & 96.9 & 141.7 & 108.2 & 78.8 & 133.7 & 108.2 & 112.9 & 110.6\\
\bottomrule
\specialrule{0em}{1.5pt}{1.5pt}
\bottomrule
\end{tabular}}
\vspace{-0.3cm}
\end{table*}

\subsection{Key Parameters Analysis}

\begin{figure}[h]
\vspace{-1em}
 \begin{center}
 \ifthenelse{\equal{\figureresolution}{low resolution}}
    {\includegraphics[angle=0,width=0.99\textwidth]{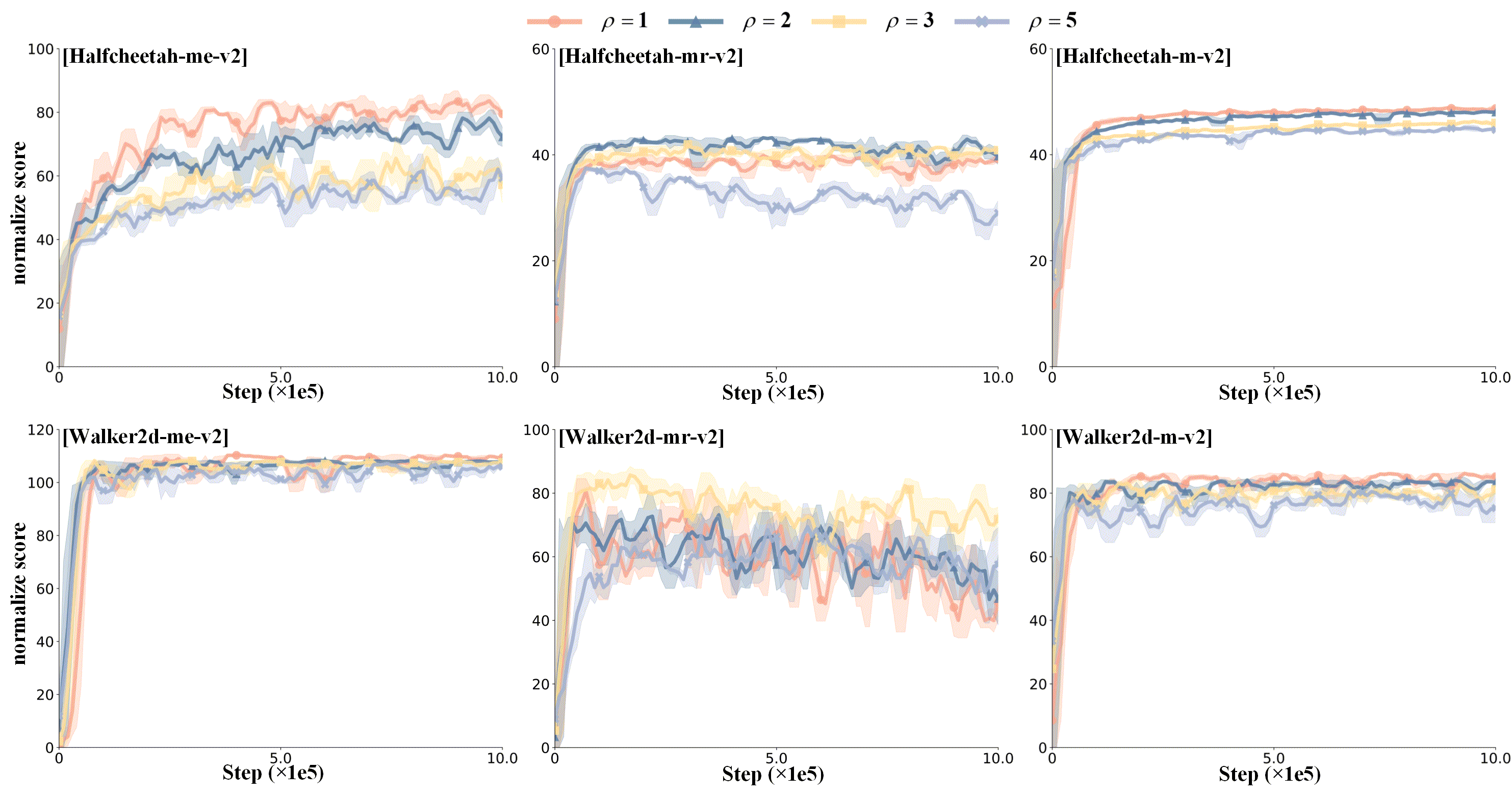}}
    {\includegraphics[angle=0,width=0.99\textwidth]{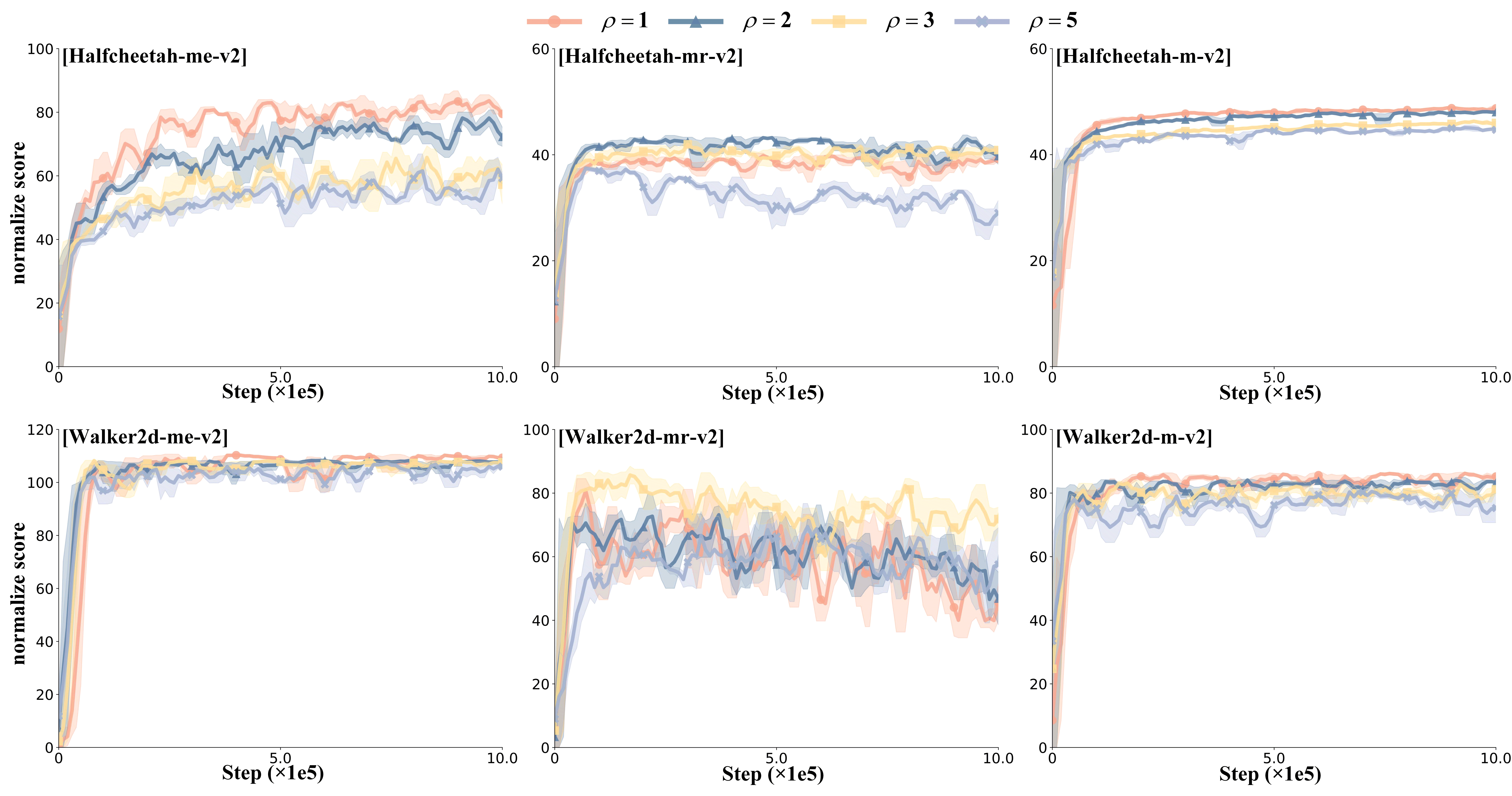}}
 \caption{Parameter sensitivity of behavior tradeoff parameter $\rho$. We investigate the influence of $\rho$ on the Gym-MuJoCo tasks.}
 \label{d4rl parameter sensitiviy on rho}
 \end{center}
 \vspace{-0.3cm}
 \end{figure}
 
We conduct experiments on key parameters $\rho$ and the generation time step $T$ of the flow model.
In Figure~\ref{d4rl parameter sensitiviy on rho}, we report the results on Gym-MuJoCo tasks, where the results show that $\rho=1$ is good for most tasks.
The results, such as sub-figures of `Halfcheetah-me-v2' and `Halfcheetah-mr-v2', show that a larger $\rho$ value will lead to tighter constraints with the behavior flow policy and result in lower performance.

\begin{figure}[t!]
\vspace{-1em}
 \begin{center}
 \ifthenelse{\equal{\figureresolution}{low resolution}}
    {\includegraphics[angle=0,width=0.99\textwidth]{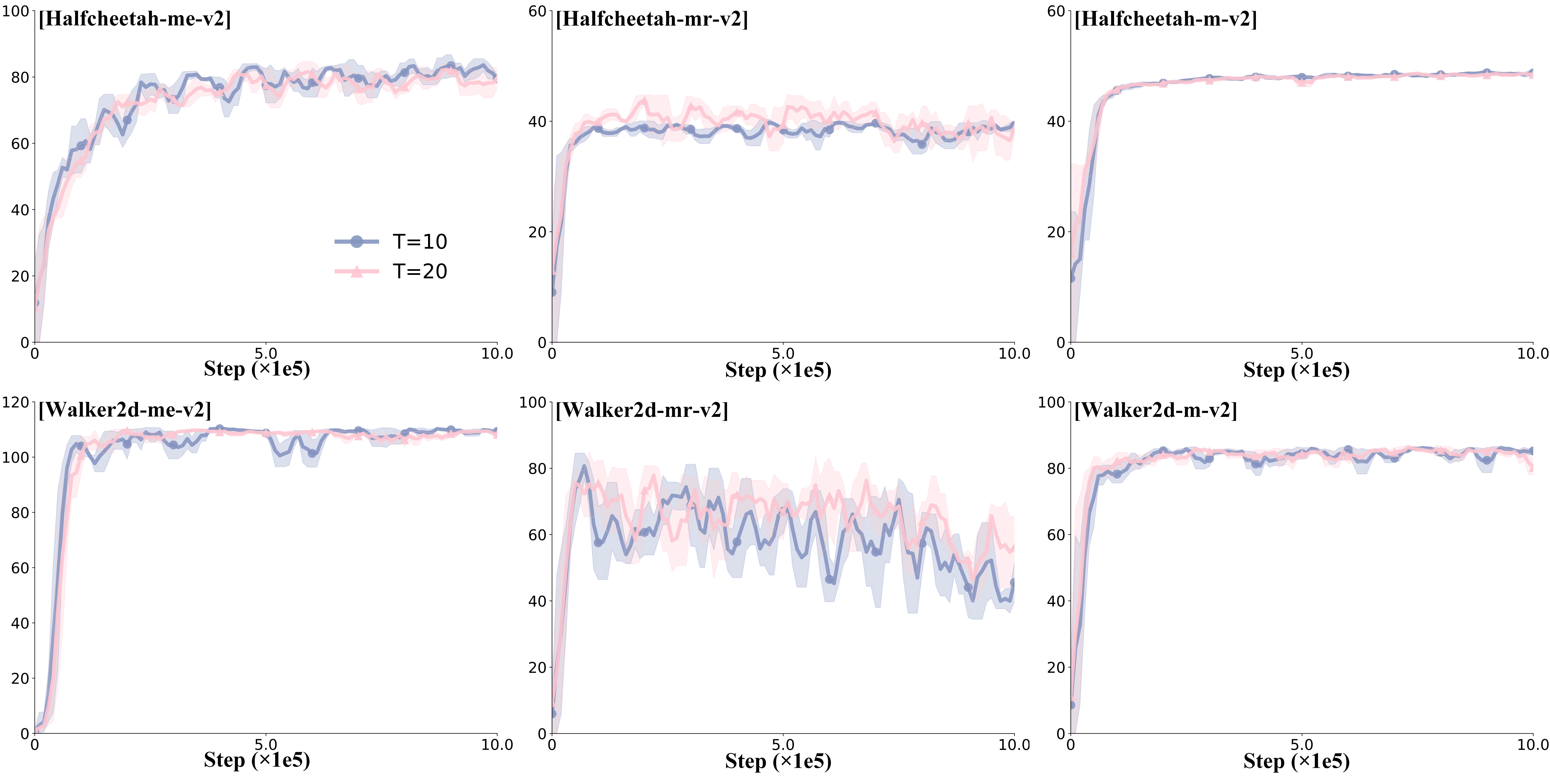}}
    {\includegraphics[angle=0,width=0.99\textwidth]{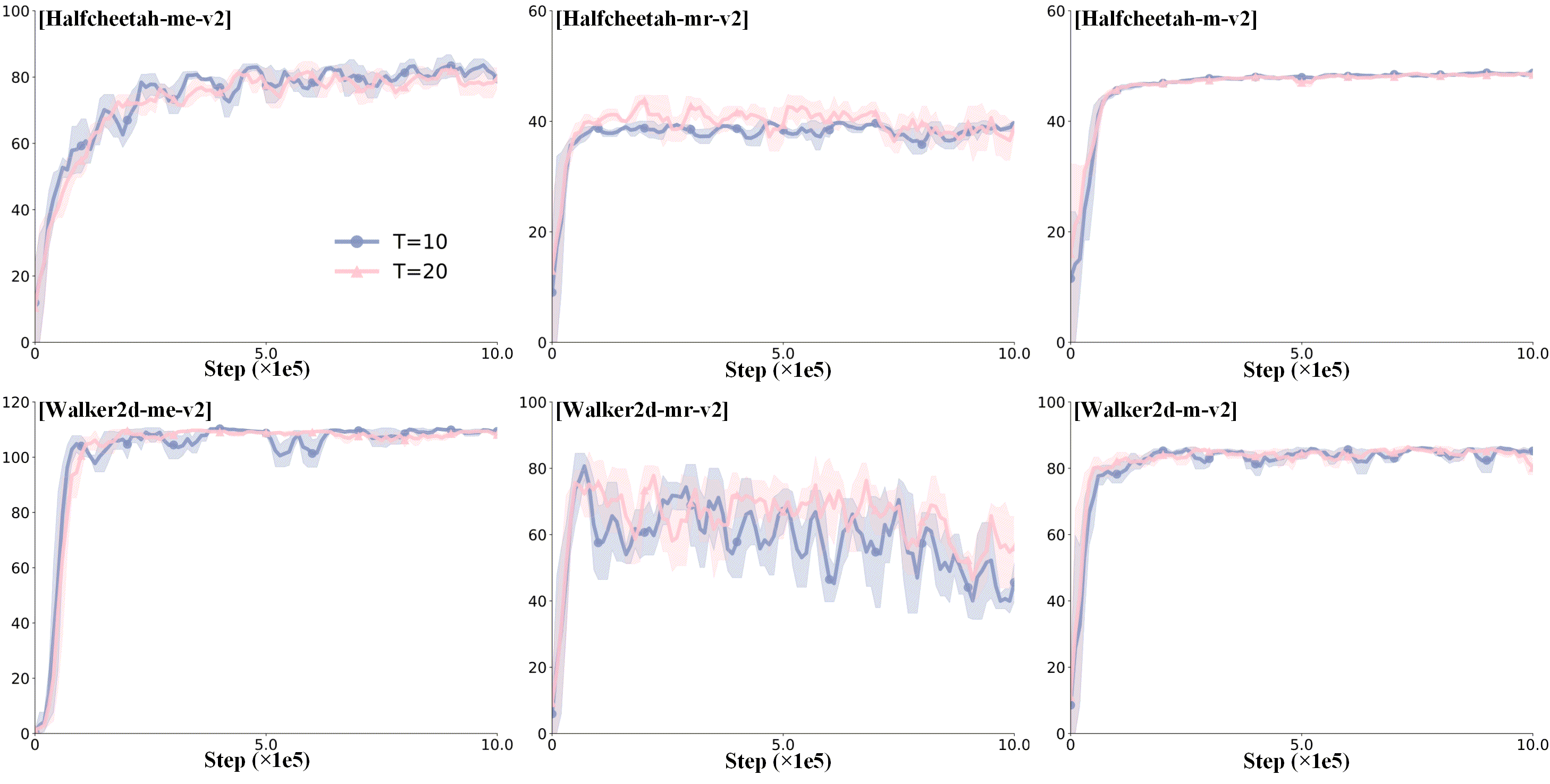}}
 \caption{Parameter sensitivity of flow time step $T$. We investigate the influence of $T$ on the Gym-MuJoCo tasks.}
 \label{d4rl parameter sensitiviy on flow time step}
 \end{center}
 \vspace{-0.3cm}
 \end{figure}

Flow time step $T$ is important for flow models to generate the actions because it directly affects the generation efficiency.
Figure~\ref{d4rl parameter sensitiviy on flow time step} shows the experiments of parameter sensitivity on flow time step $T$.
From the results, we know that a longer generation time step brings relatively smaller performance gains on tasks, such as `Halfcheetah-mr-v2' and `Walker2d-mr-v2'. 
In most tasks, $T=10$ is as good as $T=20$ and we adopt $T=10$ in the experiments of this paper.

\section{Training Details}

\subsection{Training of Traditional Q Function}\label{Training of Traditional Q Function}

In practice, we can adopt various strategies~\cite{van2016deep, kostrikov2021offline, lu2023contrastive} to learn the traditional state-action value functions.
In this paper, we adopt the strategy introduced in implicit q-learning (IQL)~\cite{kostrikov2021offline} as the training objective
\begin{align}\label{IQL V loss}
    &\mathcal{L}_{V}=\mathbb{E}_{(s^n,a^n)\sim D}\left[L_2^\tau(V_{\varphi}(s^n)-Q_{\bar{\psi}}(s^n,a^n))\right],\\ \label{IQL Q loss}
    &\mathcal{L}_{Q}=\mathbb{E}_{(s^n,a^n,s^{n+1})\sim D}\left[||r(s^n,a^n) + \gamma V_{\varphi}(s^{n+1})-Q_{\psi}(s^n,a^n)||^2_2\right],\\ \label{IQL expectile loss}
    &L_2^\tau(y)=|\tau-1(y<0)|y^2,
\end{align}
where $\varphi$ is the parameter of the state value function $V_\varphi$, $Q_{\psi}$ and $Q_{\bar{\psi}}$ are the state-action value function and the target state-action value function, the parameters $\bar{\psi}$ will be updated periodically with the exponential moving average operation, and $n$ denotes the RL time step.

\begin{table}[t!]
\centering
\caption{The hyperparameters of Decision Flow (\ourmodel{}).}
\label{our method hyper}
\begin{tabular}{l l}
\toprule
Hyperparameter & Value\\
\midrule
network backbone & MLP\\
flow value function $Q^f_\Psi$ hidden layer & 4 \\
flow value function $Q^f_\Psi$ hidden layer neuron & 256/128 \\
flow value function $V^f_\Omega$ hidden layer & 4 \\
flow value function $Q^f_\Omega$ hidden layer neuron & 256/128 \\
flow value function $V^f_\chi$ hidden layer & 3 \\
flow value function $Q^f_\chi$ hidden layer neuron & 256 \\
action value function ($Q_{\psi}$) hidden layer & 3 \\
action value function ($Q_{\psi}$) hidden layer neuron & 256 \\
state value function ($V_{\phi}$) hidden layer & 3 \\
state value function ($V_{\phi}$) hidden layer neuron & 256 \\
expectile weight $\tau$ & 0.5 \\
behavior tradeoff $\rho$ & 1.0 \\
flow discounted factor $\gamma_\tau^f$ & 1.0 \\
RL discounted factor $\gamma$ & 0.99 \\
flow time step $T$ & 10 \\

\bottomrule
\end{tabular}
\end{table}

\subsection{Computation}\label{computation}
In this paper, we conduct the experiments with NVIDIA A10 GPUs and NVIDIA GeForce RTX 3090 GPUs. 
The CPU type is Intel(R) Xeon(R) Gold 6230 CPU @ 2.10GHz. 
Each run of the experiments consumes about 48-72 hours, depending on the different offline RL tasks.

\subsection{Hyperparameters}\label{Hyperparameters}
The hyperparameters used in our method are shown in Table~\ref{our method hyper}.

\subsection{Baselines Category}

The baselines used in this paper belong to several categories.
\begin{itemize}
    \item Traditional RL methods, which include actor-critic algorithms such as AWAC~\cite{nair2020awac}, IQL~\cite{kostrikov2021offline}, AWR~\cite{peng2019advantage}, and BC, model-based algorithms, such as TAP~\cite{jiang2022efficient}, MOReL~\cite{kidambi2020morel}, MOPO~\cite{yu2020mopo}, MBOP~\cite{argenson2020model}, and COMBO~\cite{yu2021combo}, policy-constraint algorithms, such as CQL~\cite{kumar2020conservative}, TD3+BC~\cite{fujimoto2021minimalist}, BCQ~\cite{fujimoto2019off}, PBRL~\cite{bai2022pessimistic}, and BEAR~\cite{kumar2019stabilizing}.
    \item Transformer-based RL methods, including BooT~\cite{wang2022bootstrapped}, TT~\cite{janner2021offline} DT~\cite{chen2021decision}, GDT~\cite{hu2023graph} and QDT~\cite{yamagata2023q}.
    \item Diffusion-based RL methods are composed of DiffuserLite~\cite{dong2024diffuserlite}, HD-DA~\cite{chen2024simple}, IDQL~\cite{hansen2023idql}, AdaptDiffuser~\cite{liang2023adaptdiffuser}, HDMI~\cite{li2023hierarchical}, SfBC~\cite{chen2022offline}, DD~\cite{ajay2022conditional}, Diffuser~\cite{janner2022planning}, D-QL~\cite{wang2022diffusion}, TCD~\cite{hu2023instructed}, QGPO~\cite{lu2023contrastive} and LD~\cite{li2023efficient}.
    \item Flow-based RL methods contain CNF~\cite{akimov2022let}, Flow~\cite{zheng2023guided}, FQL~\cite{park2025flow}, FAWAC, FBRAC, and IFQL, where FAWAC, FBRAC, and IFQL are the variants of AWAC~\cite{nair2020awac}, DQL~\cite{wang2022diffusion}, and IDQL~\cite{hansen2023idql}.
\end{itemize}

\section{Simplification of Flow Matching}\label{Simplification of Flow Matching}

In Section~\ref{Preliminary}, we introduce the general flow matching loss
\begin{equation*}
    \mathcal{L}_{fm}=\mathbb{E}_{x_t\sim p_t(x_t),t\sim U(0,1)}[||u_{\theta}(x_t,t)-u(\phi_t(x))||^2_2],
\end{equation*}
and find that the above objective is intractable to optimize because 1) $u(x_t,t)$ is a complicated object governing the joint transformation between two high-dimensional distributions; 2) we can not obtain the data that comes from $p_t(x)$.
Fortunately, the objective can be simplified drastically when conditioning on a single target example $x_1$, which leads to the tractable conditional flow matching objective
\begin{equation*}
    \mathcal{L}_{cfm}=\mathbb{E}_{x_1\sim p_1(x_1),x_t\sim p_t(x_t|x_1),t\sim U(0,1)}\left[||u_\theta(x_t,t)-u(x_t|x_1)||_2^2\right].
\end{equation*}
Remarkably, the objectives of $\mathcal{L}_{fm}$ and $\mathcal{L}_{fm}$ provide the same gradients to learn $u_\theta$.
\begin{proof}
    \begin{align}
        \mathcal{L}_{fm}&=\mathbb{E}_{t,p_t(x_t)}\left[||u_\theta(x_t,t)-u_t(x_t)||_2^2\right]\\
            &=\mathbb{E}_{t,p_t(x_t)}\langle u_\theta(x_t,t),u_t(x_t)\rangle\\
            &=\mathbb{E}_{t,p_t(x_t)}\langle u_\theta(x_t,t),\int u_t(x_t|x_1)p_t(x_1|x_t)dx_1\rangle\\
            &=\mathbb{E}_{t,p_t(x_t)}\langle u_\theta(x_t,t),\int u_t(x_t|x_1)\frac{p_t(x_t|x_1)p(x_1)}{p_t(x_t)}dx_1\rangle\\
            &=\int_{t,p_t(x_t)}\left[u_\theta(x_t,t)*\int_{p_t(x_1|x_t)} u_t(x_t|x_1)\frac{p_t(x_t|x_1)p(x_1)}{p_t(x_t)}dx_1\right] p_t(x_t)dx_tdt\\
            &=\int_{t,p_t(x_t)}\left[u_\theta(x_t,t)*\int_{p_t(x_1|x_t)} u_t(x_t|x_1)\frac{p_t(x_t|x_1)p(x_1)}{p_t(x_t)}dx_1\right] p_t(x_t)dx_tdt\\
            &=\int_{t,p_t(x_t)}\int_{p_t(x_1|x_t)}\left[u_\theta(x_t,t)* u_t(x_t|x_1)p_t(x_t|x_1)p(x_1)dx_1\right]dx_tdt\\
            &=\mathbb{E}_{t,q(x_1),p_t(x_t|x_1)}\langle u_\theta(x_t,t), u_t(x_t|x_1)\rangle\\
            &=\mathbb{E}_{t,q(x_1),p_t(x_t|x_1)}\left[||u_\theta(x_t,t)-u(x_t|x_1)||_2^2\right]\\
            &=\mathcal{L}_{cfm}
    \end{align}
\end{proof}

\section{Proof of Direction-Oriented Decision Flow}\label{Proof of Direction-Oriented Decision Flow}

Here, we recall that the corresponding definition of flow RL (Refer to Definition~\ref{Flow MDP in mainbody} for details), where we use $n$ and $t$ to represent the RL time step and the flow time step.
Besides, we denote the traditional action value function with $Q(s,a_1)$, the flow action value function with $Q^f(s,a_t,u_t)$, and the flow value function with $V^f(s,a_t,\hat{u}_t)$, where $\hat{u}_t=\frac{u_t}{||u_t||}$ and $u_t=u_\theta(s,a_t,t)$.
Based on the Definition~\ref{Flow MDP in mainbody}, we make the following Lemmas and Theorems.

\begin{lemma}\label{Critic Consistency}
(Critic Consistency)
If $Q\rightarrow Q^*$, where $Q^*$ is the optimal conventional critic and $Q^f$ and $V^f$ with sufficient model capacity, and the objectives $\mathcal{L}_{Q^f}$ and $\mathcal{L}_{V^f}$ is defined as
\begin{align}
    \mathcal{L}_{Q^f}&=\mathbb{E}\left[||Q_{\Psi}^f-Q||_2^2\right],\\
    \mathcal{L}_{V^f}&=\mathbb{E}\left[||V_{\Omega}^f-Q||_2^2\right].
\end{align}
Then, we will conclude that
\begin{equation}
    Q^{f*}(s,a_t,u_t)=V^{f*}(s,a_t,\hat{u}_t)=Q^*(s,a_1),\forall t,a_t.
\end{equation}
\begin{proof}
    With perfect optimization, the MSE loss forces the outputs to equal the target $Q(s, a_1)$. 
    As $Q\rightarrow Q^*$, the flow values $Q^f$ and $V^f$ output the value of $Q^*$.
\end{proof}
\end{lemma}

As the convergence of flow value functions, we show that the updating signals from $Q^f$ and $V^f$ are aligned with the traditional $Q$ function, and the intermediate flow actions (velocity field) will lead to the high-return region, as shown in Lemma~\ref{Direction Optimality} and Lemma~\ref{Monotone Critic Improvement}.

\begin{lemma}\label{Direction Optimality}
(Direction Optimality)
Under the results of Lemma~\ref{Critic Consistency}, the gradient of $V^{f*}$ with respect to $\hat{u}_t$ vanishes if and only if $\hat{u}_t$ is aligned with the gradient of action $\nabla_{a_t}Q^*(s, a_t)$.

\begin{proof}
Let $F(\hat{u}_t)=V^{f*}(s,a_t,\hat{u}_t)$. 
Considering that $||\hat{u}_t||_2^2=1$, we can obtain the stationary point of the Lagrangian about $F(\hat{u}_t)$
\begin{equation}
    L(\hat{u}_t,\eta)=F(\hat{u}_t)-\eta*0.5*(||\hat{u}_t||_2^2-1).
\end{equation}
\begin{equation}
    \nabla_{\hat{u}_t} L(\hat{u}_t,\eta)=0\Rightarrow \nabla_{\hat{u}_t}F(\hat{u}_t)=\eta \hat{u}_t
\end{equation}
indicates that at any stationary point the gradient of $F(\hat{u}_t)$ is parallel to the direction $\hat{u}_t$.
Considering the first-order Taylor expansion~\cite{vetter1973matrix} of $F(\hat{u}_t)$
\begin{align}
    F(\hat{u}_t)&=Q^*(s,\Gamma(a_t+h\lambda\hat{u}_t)),\\
    &=Q^*(s,\Gamma(a_t))+h\lambda (\nabla_{a} Q^*(s,\Gamma(a_t)))^\top J_{\Gamma}(a_t)\hat{u}_t + O((h\lambda)^2),
\end{align}
where $J_{\Gamma}(a_t)$ is the Jacobian of $\Gamma$ at $a_t$, $\Gamma$ is the flow from $a_{t+\Delta t}=a_t+h\lambda\hat{u}_t$ to $a_1$ with small time step $h\lambda\hat{u}_t$.
We get the approximate gradient of $F(\hat{u}_t)$ w.r.t. $\hat{u}_t$
\begin{equation}
    \nabla_{\hat{u}_t}F(\hat{u}_t)=g_t=h\lambda (\nabla_{a_t} Q^*(s,\Gamma(a_t)))^\top J_{\Gamma}(a_t).
\end{equation}
Note that $J_{\Gamma}(a_t)=I+O(h\lambda)$ is full-rank because the flow transformation from $a_{t+\Delta t}$ to $a_1$ is smooth and time-invertible.
$J_{\Gamma}(a_t)=I+O(h\lambda)$ can be proved by decomposing the transformation from $a_t$ to $a_1$ with $\Gamma(a_t)=\Gamma(\varPhi(a_t))$~\cite{ho2019flow++}, where $a_{t+\Delta t}=\varPhi(a_t)=a_t+h\lambda\hat{u}_t$.
So the Jacobian $J_{\Gamma}(a_t)$ from $a_t$ to $a_1$ is 
\begin{equation*}
    J_{\Gamma}(a_t)=J_{\Gamma}(a_{t+\Delta t})J_{\varPhi}(a_t),
\end{equation*}
where 
\begin{equation}\label{Jacobian w.r.t. a_t}
    J_{\varPhi}(a_t)=I+h\lambda[\nabla\hat{u}](a_t)+O((h\lambda)^2)=I+O(h\lambda).
\end{equation}
Let $J(\tau)=\frac{\partial x(\tau)}{\partial a}$ and $x(\tau)$ solve 
\begin{equation*}
    \dot{x}(\tau)=u_\theta(x,\tau),x(t+\Delta t)=a,\tau\in[t+\Delta t, T],
\end{equation*}
where $\dot{x}$ is the derivative w.r.t. flow time step.
The variational equation of Jacobian $J$ is
\begin{equation*}
    \frac{d J(\tau)}{d\tau}=\nabla_{x}u_\theta(x(\tau),\tau)J(\tau),J(t+\Delta t)=I,
\end{equation*}
where we set $A(\tau)=\nabla_{x}u_\theta(x(\tau),\tau)$.
Integrating over $\Delta \tau=T-(t+\Delta t)$ yields the Peano-Baker series~\cite{baake2011peano}
\begin{equation*}
    J(T)=I+\int_{t+\Delta t}^{T}A(\tau_1)d\tau_1+\int_{t+\Delta t}^{T}A(\tau_1)A(\tau_2)d\tau_2d\tau_1+...
\end{equation*}
Assume $u_\theta$ is L-Lipschitz~\cite{heinonen2005lectures}, i.e., $||A(\tau)||<L$, we obtain
\begin{equation*}
    ||J(T)-I||\leq c_0=e^{L\Delta \tau}-1,
\end{equation*}
By the mean-value theorem~\cite{elliott2012probabilistic}, moving the starting point from $a_t$ to $a_{t+\Delta t}$ changes the Jacobian by at most a Lipschitz constant $L_J$ times $h\lambda||\hat{u}_t||$
\begin{equation*}
    ||J_{\Gamma}(a_{t+\Delta t})-J_{\Gamma}(a_{t})||\leq L_J(h\lambda).
\end{equation*}
Thus we have
\begin{equation}\label{Jacobian w.r.t. a_{t+1}}
    J_{\Gamma}(a_{t+\Delta t})=I+h\lambda B_t+O((h\lambda)^2)=I+h\lambda O(h\lambda),
\end{equation}
where $B_t\leq \frac{c_0}{h\lambda}+L_J$.
Finally, combining Equation~\eqref{Jacobian w.r.t. a_t} and Equation~\eqref{Jacobian w.r.t. a_{t+1}} leads to 
\begin{equation}\label{Jacobian from a_{t} to a_{t+Delta t}}
    J_{\Gamma}(a_t)=I+O(h\lambda).
\end{equation}
For a sufficiently small step size, the angle of $g_t$ and $\nabla_{a_t} Q^*(s,\Gamma(a_t))$ is $O(h\lambda)$.
In other words, the gradient of $F(\hat{u}_t)$ and $\nabla_{a_t} Q^*(s,\Gamma(a_t))$ remain the common direction up to $O(h\lambda)$.
\end{proof}

\end{lemma}

\begin{lemma}\label{Monotone Critic Improvement}
(Monotone Critic Improvement)
Let $a_{t+\Delta t}=a_t+h\lambda\hat{u}_t$, where $\hat{u}$ satisfies the Lemma~\ref{Direction Optimality} for small $h\lambda<\epsilon$. 
Then
\begin{equation}
    Q^*(s,a_{t+\Delta t}) \geq Q^*(s,a_t),
\end{equation}
where the equality holds only when $\nabla_{a} Q^*(s,a)=0$.

\begin{proof}
By Taylor expansion, $Q^*(s,a_{t+\Delta t})=Q^*(s,a_{t})+h\lambda||\nabla_{a_t} Q^*(s,a_{t})||+\frac{1}{2}(h\lambda)^2\hat{u}^\top_tH_{Q^*}\hat{u}_t + O((h\lambda)^3)$, we obtain that the quadratic term is bounded by
\begin{equation*}
    \left|\left|\frac{1}{2}(h\lambda)^2\hat{u}^\top_tH_{Q^*}\hat{u}_t\right|\right|\leq \frac{1}{2}(h\lambda)^2H_{max},
\end{equation*}
where the first-order term is non-negative.
Selecting $h\lambda$ below the radius, i.e., $0<h\lambda<\frac{2||\nabla_{a_t} Q^*(s,a_{t})||}{H_{max}}$ where higher-order terms dominate gives monotone improvement
\begin{equation*}
    Q^*(s,a_{t+\Delta t}) \geq Q^*(s,a_t).
\end{equation*}

\end{proof}

\end{lemma}

Finally, in Theorem~\ref{flow policy convergence}, we give a summary of the iterative improvement of the flow policy and show that the trained flow policy $u_\theta$ will converge as the flow $Q^f$ and $V^f$ functions reach optimum.

\begin{theorem}\label{flow policy convergence}
Under the assumptions that 1) the traditional $Q$ function converges to the optimal value, 2) the flow value functions $Q^f$ and $V^f$ have sufficient capacity, and the training objectives are defined in Lemma~\ref{Critic Consistency}, 3) the generation step size $h\lambda$ is small, 4) $Q$ is continuously differentiable over action space and attains a unique maximum $a^*_1=\arg\max_a Q(s,a)$ and Lemma~\ref{Critic Consistency},~\ref{Direction Optimality},and~\ref{Monotone Critic Improvement}, we conclude that the generated actions by flow policy converge to the optimal actions $a^*_1$ and the flow policy $u_\theta$ is optimal.
\end{theorem}

\begin{theorem}
(Policy Improvement with Advantage Objective)
We propose to update the flow policy parameters by ascending 
\begin{equation}\label{policy loss of adv}
    \mathcal{J}^f_\theta=\mathbb{E}_{s,t,a_t}[Q_{\Psi}^f(s,a_t,u_\theta)-V_{\Omega}^f(s,a_t,\hat{u}_\theta)].
\end{equation}
Recall from Lemma~\ref{Critic Consistency} that the difference between the flow $Q^f$ function and the flow $V^f$ function is zero at the optimum $Q^*$.
We show that ascending $\mathcal{J}^f_\theta$ is an approximate natural gradient step on the true performance
\begin{equation*}
    \mathcal{J}_\theta = \mathbb{E}_{s}[Q^*(s,a_1)].
\end{equation*}

\begin{proof}
For any intermediate flow step t, 
\begin{equation*}
    \frac{\partial a_1}{\partial \theta}=\frac{\partial a_1}{\partial a_t}\frac{\partial a_t}{\partial \theta}=J_{\Gamma}(a_t)\frac{\partial a_t}{\partial \theta}.
\end{equation*}
From Equation~\eqref{Jacobian from a_{t} to a_{t+Delta t}}, we can treat $J_{\Gamma}(a_t)$ as identity from the first order, then
\begin{equation}
    \nabla_\theta \mathcal{J}_\theta=\mathbb{E}\left[\nabla_{a_1} Q^*(s,a_1)\frac{\partial a_1}{\partial \theta}\right]\approx \mathbb{E}\left[\nabla_{a_t} Q^*(s,a_t)\frac{\partial a_t}{\partial \theta}\right].
\end{equation}
Expanding $Q^*(s,\cdot)$ at $a_t$
\begin{equation*}
    Q^*(s,\cdot) = Q^*(s,a_t)+h\lambda\langle\nabla_a Q^*(s,a_t),\hat{u}\rangle + O((h\lambda)^2).
\end{equation*}
Noted that $V^f$ is independent of $\lambda$ because $V^f$ just measure the expected return on the direction of $u_{\theta}$ and $Q^f$ is dependent of $\lambda$, then we have
\begin{align}
    V^{f*} &= Q^*(s,a_t) + O((h\lambda)^2),\\
    Q^{f*} &= Q^*(s,a_t) + h\lambda\nabla_a Q^*(s,a_t)\hat{u} + O((h\lambda)^2),
\end{align}
which indicates that 
\begin{equation}
    \nabla_\theta\mathcal{J}^{f*}_\theta = \nabla_\theta \mathbb{E}[Q^{f*} - V^{f*}]\approx \mathbb{E}\left[h\lambda \nabla_{a_t} Q^*(s, a_t)\frac{\partial a_t}{\partial \theta}\right].
\end{equation}
Hence, ascending the surrogate objective Equation~\eqref{policy loss of adv} performs stochastic gradient ascent on the optimal Q value function under first order.

\end{proof}

\end{theorem}

\section{Proof of Divergence-Oriented Decision Flow}\label{Proof of Divergence-Oriented Multi-step Decision Flow}

\begin{lemma}
For continuous actions, the Bellman operator
\begin{equation*}
    (\mathcal{T}Q)(s,a)=\mathbb{E}_{s^\prime}\left[r+\gamma\max_{a^\prime}Q(s^\prime, a^\prime)\right]
\end{equation*}
is $\gamma$-contraction and it guarantees that the parameterized Q function of RL has a single globally asymptotically stable equilibrium at $Q^*$.

\end{lemma}

\begin{definition}\label{Flow MDP definition}
(Flow MDP for \ourmodel{}-div)
We regard the generation process induced by the velocity field as Flow-MDP, where the flow state at the flow time step t is $(a^{n-1},s^n,a_t^n)$, the flow action is $a_{t+\Delta t}^n$ or $a_{\tau+1}$, $\tau\in\{0,...,T\}$, $t\in[0,1]$, $\Delta t=\frac{1}{T}$, $t=\tau*\Delta t$, $T$ is the predefined total flow time steps for each action generation, the reward is defined as
\begin{equation}\label{reward aggregation r}
    r_\tau^{flow}=
        \begin{cases}
            -\mathcal{D}(u_{\theta}||u_v), &0\leq \tau < T,\\
            Q(s^n,a_1^n), &\tau=T,
    \end{cases}
\end{equation}
$u_{\theta}$ is the flow policy, $\theta$ is the parameters of the flow policy, $u_v$ is the behavior flow policy, the discounted factor $\gamma^f=1$ for all flow time steps, $\mathcal{D}(u_{\theta}||u_v)$ is the divergence between $u_\theta$ and $u_v$, and we adopt $\mathcal{D}(u_\theta(s,a_t,t)||u_v(s,a_t,t))=||u_\theta - u_\upsilon||_2$ in practice.
 
Mathematically, the $T$ step flow decision policy optimization problem is defined as
\begin{equation}
    \max \mathbb{E}\left[\sum_{\tau=0}^{T}(\gamma^f)^\tau r_\tau^{flow}\right].
\end{equation}
The flow MDP transition is defined as
\begin{align}
    \mathcal{T}_V^f V^f_{\chi}(a^{n-1},s^n,a_1^{n})&=Q(s,a_1),\\
    \mathcal{T}_V^f V^f_{\chi}(a^{n-1},s^n,a_t^{n})&=-\mathcal{D}(u_\theta||u_\upsilon)+V^f_{\chi}(a^{n-1},s^n,a_{t+\Delta t}^{n}),
\end{align}
where $\mathcal{T}_V^f V_{\chi}(a^{n-1},s^n,a_t^{n})$ represents the Bellman operator on the intermediate flow steps, $V^f_{\chi}$ is the flow value function, $\chi$ denotes the parameters of the flow value function $V^{f}_{\chi}$, and $a^{n-1}$ is the actions at the last time step, which facilitates stable training of $V^f_{\chi}$.

The optimal flow value function of the flow MDP satisfies
\begin{equation}
    V^{f*}_{\chi}(a^{n-1},s^n,a_t^n)=-\mathcal{D}(u_{\theta}||u_v)+V^{f*}_{\chi}(a^{n-1},s^n,a_{t+\Delta t}^n), 
\end{equation}
where the terminal condition $V^{f*}_{\chi}(a^{n-1},s^n,a_1^n)=Q^*(s^n,a_1^n)$.
Mathematically, inspired by the definition of traditional Bellman error~\cite{sutton1998reinforcement}, training the flow value function is equal to minimizing  
\begin{equation}\label{mse loss of V_chi}
    \mathcal{L}_{V_\chi}^{f}=\frac{1}{2}\mathbb{E}_{a^{n-1},s^n,a^{n}, t\sim U(0,1),a^{n}_t=\phi_t(a^{n}_t|a^{n})}\left[V_{\chi}^f(a^{n-1},s^n,a_t^n)-\hat{g}_{V^f}\right],
\end{equation}
where 
\begin{equation*}
    \hat{g}_{V^f}=-\sum_{\tau=t/\Delta t}^{T-1}\mathcal{D}(u_{\theta}(s,a_{\tau*\Delta t}, \tau*\Delta t)||u_v(s,a_{\tau*\Delta t}, \tau*\Delta t))+Q(s,a_1)
\end{equation*}
is the estimation of $V^{f*}_{\chi}(a^{n-1},s^n,a_t^n)$.

\end{definition}

\begin{lemma}\label{Flow Policy Evaluation}
(Flow Policy Evaluation)
Consider that the flow bellman backup operator $\mathcal{T}^f_V$ that makes
\begin{equation*}
    \mathcal{T}^f_VV^{f}_{\chi}(a^{n-1},s^n,a_t^n)=-\mathcal{D}(u_{\theta}||u_v)+V^{f}_{\chi}(a^{n-1},s^n,a_{t+\Delta t}^n),
\end{equation*}
flow time step $T<\infty$, and $r_t^{flow}$ is bounded by a constant $r_{max}^{flow}$.
Under the Definition~\ref{Flow MDP definition}, then the flow value function will converge to the unique optimal value function $V_\chi^{f*}$ with the convergence of $Q$.

\begin{proof}
The mean-square error Equation~\eqref{mse loss of V_chi} is a strictly convex functional. Given sufficient capacity of $V_\chi^f$, we can apply the standard convergence results of policy evaluation shown in~\cite{sutton1998reinforcement, haarnoja2018soft} to prove the results.
\end{proof}

\end{lemma}

\begin{lemma}\label{Flow policy Improvement}
(Flow Policy Improvement)
Let $u_\theta^{new}$ be the optimizer that maximizes 
\begin{equation}
    \mathcal{J}_{u_\theta} = -\mathcal{D}(u_{\theta}||u_v)+V^{f}_{\chi}(a^{n-1},s^n,a_{t+\Delta t}^n).
\end{equation}
Then $\mathcal{J}_{u_\theta}^{new}(a_t^n)\geq\mathcal{J}_{u_\theta}^{old}(a_t^n)$ for all $(a^{n-1}, s^{n},a_t^{n})$.

\begin{proof}
The optimizer $u_\theta^{new}$ that can maximize $-\mathcal{D}(u_{\theta}||u_v)+V^{f}_{\chi}(a^{n-1},s^n,a_{t+\Delta t}^n)$ is guaranteed to satisfy
\begin{equation*}
-\mathcal{D}(u_{\theta}^{new}||u_v)+V^{f}_{\chi_{old}}(a^{n-1},s^n,a_{t+\Delta t}^n)\geq-\mathcal{D}(u_{\theta}^{old}||u_v)+V^{f}_{\chi_{old}}(a^{n-1},s^n,a_{t+\Delta t}^n).
\end{equation*}
Applying follow bellman backup operator $\mathcal{T}_V^f$ recursively, we will obtain
\begin{align*}
\mathcal{J}_{u_\theta}^{old}(a_t^n)&\leq -\mathcal{D}(u_{\theta}^{new}(s^n,a_t^n,t)||u_v(s^n,a_t^n,t))+V^{f}_{\chi_{old}}(a^{n-1},s^n,a_{t+\Delta t}^n)\\
&\leq -\mathcal{D}(u_{\theta}^{new}(s^n,a_t^n,t)||u_v(s^n,a_t^n,t))\\
&~+\left[-\mathcal{D}(u_{\theta}^{new}(s^n,a_{t+\Delta t}^n,t+\Delta t)||u_v(s^n,a_{t+\Delta t}^n,t+\Delta t))+V^{f}_{\chi_{old}}(a^{n-1},s^n,a_{t+2\Delta t}^n)\right]\\
&\leq ...\\
&\leq \mathcal{J}_{u_\theta}^{new}(a_t^n).
\end{align*}
Noted that during the training of the flow value function, the Q function is fixed, which means that $Q_\psi$ is same for $\mathcal{J}_{u_\theta}^{new}(a_t^n)$ and $\mathcal{J}_{u_\theta}^{old}(a_t^n)$.

\end{proof}

\end{lemma}

\begin{theorem}
(Flow Policy Iteration)
Repeated application of flow policy evaluation (Lemma~\ref{Flow Policy Evaluation}) and flow policy improvement (Lemma~\ref{Flow policy Improvement}) to any flow policy $u\in\mathcal{U}$ converges to a flow policy $u_\theta^*$ such that $\mathcal{J}_{u^*_\theta}(a_t^n)\geq\mathcal{J}_{u_\theta}(a_t^n)$ for all $u\in\mathcal{U}$ and $(a^{n-1},s^n,a^n_t)\in{\mathcal{A}\times\mathcal{S}\times\mathcal{A}}$.

\begin{proof}
The proof can be obtained from Lemma~\ref{Flow Policy Evaluation} and Lemma~\ref{Flow policy Improvement}.
\end{proof}
\end{theorem}

\end{document}